\pdfoutput=1

\documentclass[11pt]{article}

\usepackage[final]{acl}

\usepackage{times}
\usepackage{latexsym}

\usepackage[T1]{fontenc}

\usepackage[utf8]{inputenc}

\usepackage{microtype}

\usepackage{inconsolata}

\usepackage{graphicx}
\usepackage{enumitem}
\usepackage{booktabs}
\usepackage{amsmath}
\usepackage{hyperref}
\usepackage{tablefootnote}
\usepackage{listings}
\title{HMT: Hierarchical Memory Transformer for Efficient Long Context Language Processing}

\author{
    Zifan He \textsuperscript{1}, Yingqi Cao \textsuperscript{2}, Zongyue Qin \textsuperscript{1}, Neha Prakriya \textsuperscript{1}, \\ \textbf{Yizhou Sun \textsuperscript{1}}, \textbf{and Jason Cong \textsuperscript{1}}
    \\
    \textsuperscript{1} University of California, Los Angeles,
    \textsuperscript{2} University of California, San Diego
    \\
    \texttt{zifanhe1202@g.ucla.edu, yic033@ucsd.edu, } \\
    \texttt{\{qinzongyue, nehaprakriya, yzsun, cong\}@cs.ucla.edu}
}

\begin{document}
\maketitle
\begin{abstract}
Transformer-based large language models (LLM) have been widely used in language processing applications. However, due to the memory constraints of the devices, most of them restrict the context window. Even though recurrent models in previous works can memorize past tokens to enable unlimited context and maintain effectiveness, they have ``flat'' memory architectures. Such architectures have limitations in selecting and filtering information. Since humans are good at learning and self-adjustment, we believe that imitating brain memory hierarchy is beneficial for model memorization. Thus, we propose the Hierarchical Memory Transformer (HMT) \footnote{\url{https://github.com/OswaldHe/HMT-pytorch}}, a novel framework that facilitates a model's long-context processing ability by imitating human memorization behavior. Leveraging memory-augmented segment-level recurrence, we organize the memory hierarchy by preserving tokens from early input segments, passing memory embeddings along the sequence, and recalling relevant information from history. Evaluating general language modeling, question-answering tasks, and the summarization task, we show that HMT consistently improves the long-context processing ability of existing models. Furthermore, HMT achieves a comparable or superior generation quality to long-context LLMs with $2 \sim 57\times$ fewer parameters and $2.5 \sim 116\times$ less inference memory, significantly outperforming previous memory-augmented models. 

\end{abstract}

\section{Introduction}
Transformer \cite{vaswani2017attention} has demonstrated its strength in contextual learning and is utilized in various applications in language processing \cite{dong2019unified} and computer vision \cite{dosovitskiy2020image}. For a decoder-only transformer model, each transformer block contains a self-attention and a feedforward network module. An optimized self-attention layer has a quadratic computational and linear space complexity \cite{dao2022flashattention} regarding the sequence length since it computes interactions between each token and all previous tokens in the sequence. To maintain the inference speed and satisfy memory requirements, most transformer models enforce maximum sequence length. For example, the Llama 3 model is designed to process 8192 tokens \cite{dubey2024llama} and the Llama 2 can process up to 4096 tokens \cite{touvron2023llama}. However, real-world applications involving long documents, such as book summarization \cite{rae2019compressive} and lifelong question-answering tasks \cite{sun2019lamol, dai2022lifelong}, can have an enormous or even infinite stream of inputs. 

Existing research attempts to build long context transformers using sparse attention \cite{beltagy2020longformer, zhang2021poolingformer, kitaev2020reformer}, retrieval-augmented models \cite{bertsch2023unlimiformer, wu2022memorizing}, and recurrent sequence models \cite{peng2023rwkv,gu2023mamba,rae2019compressive}. Still, these models face at least one of two issues: (1) difficulty in adapting to future models due to a change in the core model architecture and (2) low effectiveness for long-range inputs under frequent context switching. In this work, we propose the Hierarchical Memory Transformer (HMT), a novel framework to enable and augment models' long-context processing ability. HMT transforms models into a memory-augmented recurrent model that imitates the brain's memory hierarchy and human memorization behavior. It has the following unique features:

\textbf{Hierarchical Memorization:} HMT mimics the memory hierarchy of the brain \cite{burgin2011epistemic} employing both learned memory tokens and current input tokens. HMT stratifies memory into sensory, short-term, and long-term, with interactions between each other.

\textbf{Memory Retrieval Mechanism:} HMT imitates memory recall by storing encoded memory embeddings generated from previous iterations and searching based on the relevance to current token segments. 

One key advantage of utilizing HMT over other memory-augment models is that HMT is a \textbf{model-independent plug-and-play} framework: future decoder-only models can directly serve as the backbone model of HMT to augment their long context processing ability without extra implementation efforts. With joint training and fine-tuning of newly introduced and original parameters of the backbone model, HMT is applicable to a wide range of LLMs, including transformer-based models and state-space models. Our contributions include: 

\begin{itemize}[wide=0pt]
    \item \textbf{HMT consistently improves models' generation quality with long context for various model architectures.} We demonstrate HMT on both transformer-based architecture and state-space models. Evaluating on Wikitext-103, PG-19 \cite{rae2019compressive}, and PubMedQA \cite{jin2019pubmedqa} datasets with multiple contexts concatenated, HMT can improve the effectiveness by up to 25.5\% in perplexity and 1.0\% higher prediction accuracy over the baseline models.
    
    \item \textbf{HMT with small backbone models can outperform large models trained on longer context samples, implying a high memory efficiency.} We evaluate HMT with SmolLM \cite{allal2024SmolLM}, OPT \cite{zhang2022opt}, and OpenLlamaV2 \cite{openlm2023openllama} models on the LongBench \cite{bai2023longbench} benchmark. In sum, HMT can achieve comparable or higher metric results with $2 \sim 57\times$ fewer parameters and $2.5 \sim 116 \times$ lower inference memory requirement than long-context large language models.

    \item \textbf{HMT surpasses previous methods specialized for efficient long-context processing by compressing contexts.} We compare HMT with RMT \cite{bulatov2022recurrent}, LongMem \cite{wang2024augmenting}, Memorizing Transformer \cite{wu2022memorizing}, CCM \cite{kim2023compressed}, and HOMER \cite{song2024hierarchical}, which are recent SoTA of memory-augmented and hierarchical methods. With the same or similar size backbone model, HMT has a better generation quality in both general language modeling and QA tasks. Furthermore, HMT has a lower memory complexity, indicating better scalability as the input length increases.
\end{itemize}

\section{Related Works and Problem Formulation} 


We will first discuss the existing efforts on long-range transformers and recurrent sequence models for infinitely long context language processing. Then, we highlight a problem that is crucial in real-world applications.

\subsection{Long Context Transformers} 
Since one of the bottlenecks of transformers is the quadratic computational complexity of self-attention, a natural approach is sparsifying attention computation. A naive sparse attention pattern is the sliding window attention \cite{kovaleva2019revealing}, where each token attends to neighbors within a local window. However, this neglects long-range interaction between words. Existing works such as Longformer \cite{beltagy2020longformer} and Poolingformer \cite{zhang2021poolingformer} extend the sliding window attention by adding global attending tokens and applying pooling to expand the receptive field area. Unlimiformer \cite{bertsch2023unlimiformer} adopts the retrieval-augmented generative method by searching the top K most important tokens for the incoming sequence. It then applies attention to just those tokens in the decoders, resulting in pruned computations with minor losses. Nevertheless, the contribution of less relevant tokens may accumulate over time and impact the overall sequence generation. Although these methods extend the attainable context length, they cannot prevent increasing memory consumption as the input length increases. Alternatively, compressing past tokens using a recurrent sequence model can potentially reduce memory consumption by condensing the information into a fixed-size embedding.

\subsection{Recurrent Sequence Models} 
Recurrent Neural Networks (RNN) have been extensively explored in sequence processing research, including Long Short-term Memory \cite{hochreiter1997long} and Gated Recurrent Unit \cite{chung2014empirical}. They reveal that RNNs perform well in memorizing past information and are hardware-friendly for implementing customized accelerators \cite{chang2015recurrent}. However, RNNs have limited advantages in learning contextual relationships between words compared with self-attention in language processing \cite{bahdanau2014neural}. One approach to alleviate this issue is the coarse-grain recurrence, in which the model splits inputs into segments, performs attention inside each segment, and propagates states (i.e., compressed information as embeddings) between segments. The Compressive Transformer \cite{rae2019compressive} further stores and compresses previous states to enhance memorization. The Recurrent Memory Transformer (RMT) \cite{bulatov2022recurrent} utilizes a memory token to summarize and propagate segment information without modifying the transformer block architecture. Theoretically, they can process unlimited long sequences, but previous information will be diluted after multiple summarizations and generation quality can drop when less relevant information occupies the memory. Recent works \cite{chevalier2023adapting, kim2023compressed} aim to further optimize RMT to improve generation quality by concatenating the results of summarizations, but this sacrifices the inference memory efficiency.

Another approach augments RNN by involving interactions between the current inputs and the previous states to learn contextual relationships in a similar way as self-attention and accelerate the computation with linear convolution. One of the representatives, RWKV \cite{peng2023rwkv}, is an RNN model inspired by the attention-free transformer (AFT) \cite{zhai2021attention}. It includes a time-mixing module to learn from previous states and a channel-mixing module to learn from the previous output. Mamba \cite{gu2023mamba} is another recurrent method based on the state-space model that employs gated convolution to accelerate model inference. These models are energy and memory-efficient with fast training speed and are able to achieve high performance in memorization tasks (e.g., associative recall), but have limitations on capturing contextual relationships and filtering irrelevant information. Recent works combine transformers with Mamba \cite{lieber2024jamba, team2024jamba} to mitigate this issue, but this reintroduces the scaling issue of the transformers. 


\subsection{Problem Formulation: Adaptive Long-context Processing} 

\begin{figure*}[ht]
    \centering
    \includegraphics[width=0.9\textwidth]{ 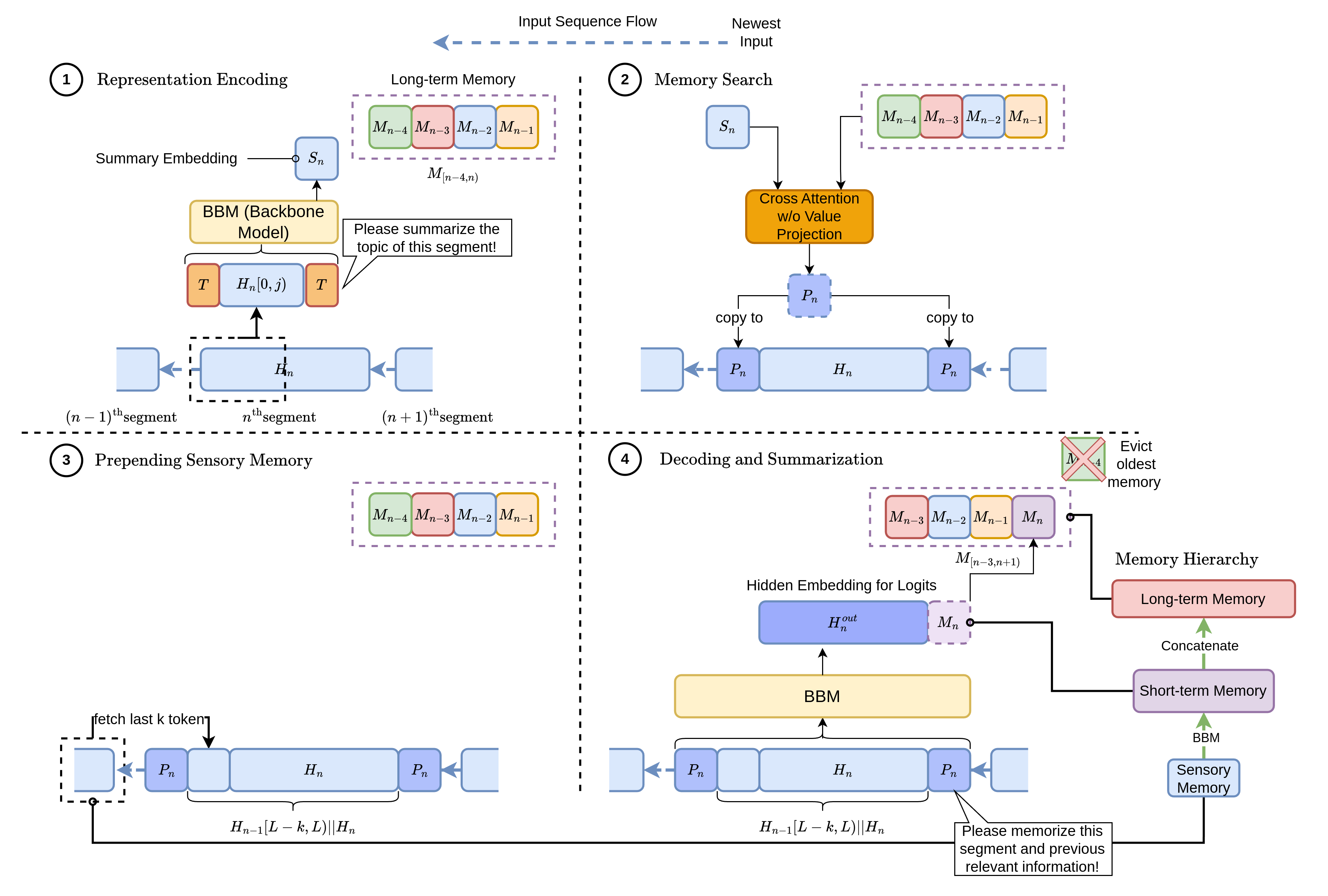}
    \caption{Overall workflow of HMT. For a segment, (1) HMT will first perform \textbf{representation encoding}, utilizing the segment summarization prompt embedding ($T$) to summarize part of the segment. (2) The generated segment summary embedding ($S_n$) is used with the cached memory embeddings for \textbf{memory search} with cross attention. The output is a memorization prompt embedding ($P_n$) which contains information relevant to the current segment. (3) The memorization prompt embedding and the last $k$ embeddings from the previous segment will augment the segment. (4) The backbone model (BBM) will process the augmented segment and generate hidden embeddings for logits ($H_n^{out}$) and the memory embedding ($M_n$), which will be pushed into the long-term memory.}
    \label{fig:recall}
\end{figure*}

We intend to develop a model that can handle \textit{infinitely long context} inputs with \textit{context adaptability}: Based on the context/topic of the input stream, the model can adaptively select past relevant information to enhance effectiveness, since irrelevant context can distract the model \cite{shi2023large}. 

In real-world applications, restrained by memory bandwidth and capacity, as well as data generation speed, long documents cannot be read as a whole by the computing hardware \cite{agerri2015big}. Furthermore, users who are constantly interacting with the language model can refer to the previous topic or switch to another topic that has high relevance to past information. For effectiveness, most recurrent models need to encode all previous inputs in the states, which can contain irrelevant information and degrade the model's quality.  

\section{Hierarchical Memory Transformer} \label{sec:method}

The main idea of HMT is to store information hierarchically and search for relevant information throughout the memory hierarchy. Table \ref{tab:symbol} describes all notations we use to illustrate the HMT architecture in this section. 



\subsection{Overall Workflow}

Given a backbone model to enhance, HMT chunks the input into $L$-token segments and operates on the hidden embeddings of the token segments ($\{H_n\}_{n=0}^\infty$), generated by the token embedding layer of the backbone model. For every segment $n$, HMT walks through four steps shown in Figure \ref{fig:recall}: 

\begin{table}
\caption{Notation used to illustrate HMT's architecture in Section \ref{sec:method} and Figure \ref{fig:recall}.}
\label{tab:symbol}
\begin{center}
\begin{small}
\begin{sc}
\resizebox{0.9\columnwidth}{!}{
\begin{tabular}{lcr}
\toprule
Notation & Meaning \\
\midrule
$H_n$ & Hidden embeddings of the $n^{th}$ segment \\
$L$ & Segment length \\
$H_n[L-k, L)$ & Last $k$ embeddings of $H_n$ \\
$H_n[0, j)$ & First $j$ embeddings of $H_n$ \\
$\text{BBM}(\cdot)$ & Backbone model \\
$P_n$ & Memorization prompt embedding \\
$H_n^{out}$ & Hidden embedding for Logits Generation \\
$M_n$ & memory embedding of the $n^{th}$ segment \\
$N$ & Number of cached memory embeddings \\
$M_{[n-N+1, n)}$ & Cached memory embeddings \\
$T$ & Segment summarization prompt embedding \\
$S_n$ & Summary embedding of the $n^{th}$ segment \\
\bottomrule
\end{tabular}
}
\end{sc}
\end{small}
\end{center}
\vskip -0.1in
\end{table}

\begin{itemize}[wide=0pt]
    \item[1)] \textit{Representation encoding} by the backbone model, which encodes part of the segment containing the essence of the ongoing topic into a single embedding to represent its context, denoted by $H_n$. 
    \item[2)] \textit{Memory search}, which utilizes the current context as a query to find relevant information in the memory.
    
    \item[3)] \textit{Prepending sensory memory}, which augments the segment to capture information in the previous segment and other relevant information. 
    
    \item[4)] \textit{Decoding and summarization}, which processes the augmented segment to get hidden embeddings for generating logits and a memory embedding that summarizes the augmented segment. 
    
\end{itemize}

The first two steps are the \textit{memory retrieval mechanism} discussed in Section \ref{sec:recall}. Steps 3 and 4 are explained in Section \ref{sec:hier_mem} along with the concept of hierarchical memorization.

\subsection{Memory Retrieval Mechanism} \label{sec:recall}


To handle context switching and prevent the intervention of irrelevant context, HMT performs memory retrieval to extract only relevant information from past knowledge. The memory retrieval mechanism involves three steps: representation extraction, memory search, and memory augmentation.

\textbf{Representation Encoding}: Depicted in Step 1 of Figure \ref{fig:recall}, HMT selects the first $j$ embeddings from the hidden embeddings of the $n^{th}$ segment, $H_n$, to extract the topic of the segment. The embeddings are augmented with the \textbf{segment summarization prompt embedding $\mathbf{T}$}. $T$ is a learnable parameter embedding, deployed to prompt the backbone model (BBM) to summarize the segment by soft prompt tuning \cite{liu2023pre}. Instead of extracting from the token embedding of BBM, we make $T$ learnable to allow a larger prompt embedding space for summarization. The backbone model will then process the augmented embeddings and generate a new embedding at the end of the output as the representation of the segment:
\begin{equation}
   S_n = \text{BBM}([T || H_n[0,j) || T])[j, j+1)
\end{equation}
where $S_n$ is the summary embedding of the $n^{th}$ segment only, $\text{BBM}(\cdot)$ is the backbone model, and ``$||$" is the concatenation operator. $S_n$ will be used for memory search.

\textbf{Memory Search}: Shown in Step 2 of Figure \ref{fig:recall}, $S_n$ is utilized as a query to find relevant memory embeddings generated from Step 4 when processing previous segments. We keep a sliding window of $N$ embeddings ($M_{[n-N+1, n)}$) and then compute:
\begin{equation}
    Q_n = S_n W_q, K_n = M_{[n-N+1, n)}W_k
\end{equation}
\begin{equation}
    P_n = \text{softmax}(\frac{Q_nK_n^T}{\sqrt{d_h}})M_{[n-N+1, n)}
\end{equation}
where $d_h$ is the hidden dimension of the cross attention. The computation is similar to cross-attention without value and output projection. $\text{Softmax}(\frac{Q_nK_n^T}{\sqrt{d_h}})$ calculates the normalized similarity score and applies it directly to $M_{[n-N+1, n)}$ to ensure similar distributions of output value and old memory tokens. We expect that the projection $W_q$ and $W_k$ can be trained such that summarizations containing similar contexts have high attention scores after projections. 

The output of a memory search is a \textbf{memorization prompt embedding} $P_n$ containing information relevant to the $n^{th}$ segment. It will be applied to augment the $n^{th}$ segment. Notice that HMT's memory is accumulative: the $n^{th}$ memory embedding contains information of all previous $n-1$ segments, with a higher loss of information for older segments. We hope that retrieving memory will strengthen the relevant memory and reduce this loss.

In practice, representation encoding is executed in parallel with the model inference on GPUs since they are independent tasks. Memory search has time complexity $O(N)$, and can also run in parallel with the segment inference when $N$ is small (e.g., $N = 300$). Thus, the overall runtime overhead of HMT is negligible.

\subsection{Hierarchical Memorization} \label{sec:hier_mem}

Human memory can be categorized into three strata: sensory memory, short-term memory, and long-term memory \cite{burgin2011epistemic}. Sensory memory refers to very short-term memory generated from sensory information, such as vision and hearing. Short-term and long-term memory are long-lasting memories, differentiated by how long they persist in the brain. HMT is inspired by this memory hierarchy. 

\textbf{Sensory Memory}: Sensory memory for the $n^{th}$ segment refers to the last $k$ token embeddings of $H_{n-1}$, $H_{n-1}[L-k, L)$. When inferencing the $n^{th}$ segment, HMT will augment the corresponding token embeddings $H_{n}$ by prepending it with $H_n[L-k, L)$, shown in Step 3 of Figure \ref{fig:recall}.  


\textbf{Short-term Memory}: HMT will encode the segment into an embedding that serves as a ``summarization" of the segment. First, HMT will append and prepend the memorization prompt embedding $P_n$ to the augmented segment. This guides the backbone model to compress the segment and relevant context into a summarization embedding with awareness of the relative positions of contexts. As depicted in Step 4 of Figure \ref{fig:recall}, we train HMT such that 
\begin{equation}
    H = \text{BBM}(P_n || H_{n-1}[L-k, L) || H_n || P_n)
\end{equation}
\begin{equation}
    H^{out}_n || M_n = H[k+1, L+k+2)
\end{equation}
where $M_n$ is the memory embedding of the $n^{th}$ segment. $H^{out}_n$ is a collection of $L$ hidden embeddings that will be used to generate logits.

\textbf{Long-term Memory}: Each generated memory embedding will be cached as the long-term memory. The cached embeddings will be utilized as the input of the \textit{memory retrieval mechanism} to generate the memorization token embedding $P_n$ for each segment as illustrated in the previous sections.

\section{Experiment} \label{sec:experiment}

We benchmark HMT with a variety of backbone models including SmolLM 135M \cite{allal2024SmolLM}, OPT 350M, OPT 2.7B \cite{zhang2022opt}, OpenLlamaV2 3B \cite{openlm2023openllama}, RWKV 3B \cite{peng2023rwkv}, and Llama 2 7B \cite{touvron2023llama}, under the same memory constraint (i.e. same maximum context window). Moreover, we test several models targeting long contexts (Mamba 370M \cite{gu2023mamba}, Yi-6B-200K \cite{young2024yi}, and Mistral 7B \cite{jiang2023mistral}) to demonstrate the benefit HMT has on generation quality and memory consumption. We evaluate HMT with state-space models (RWKV and Mamba) as backbones since we believe that models which can already process infinitely long inputs would benefit even further from HMT. All models mentioned are trained and assessed on 4 \textbf{AMD MI210 GPUs}, which can handle models up to 7B parameters. We further test HMT on 4 NVIDIA A100-80GB GPUs for the Qwen 2.5 14B model \cite{bai2023qwen} to justify its scalability to larger models and gain a consistent effectiveness boost. To tune the extra parameters introduced by HMT, we use the RedPajamaV2 \cite{together2023redpajama} dataset to pre-train each model. Notice that HMT introduced new model hyperparameters on top of the backbone model ($L$, $j$, $N$, and $k$). A common configuration is $L=1024$, $j=512$, $N=300$, and $k=32$, and we adjust these values for each model to achieve the best performance. To compare with previous works (RMT, LongMem, Memorizing Transformer, CCM), we apply the same backbone models if the method is applicable to any model, or find a backbone model with a similar size if the method requires special architecture.

For the long-context benchmark, we select subsets (NarrativeQA, Qasper, and MultiFieldQA-en for single document QA; HotpotQA, 2WikiMQA, and MuSiQue for multi-document QA; GovReport, QMSum, and Multi-News for summarization; TriviaQA for few-shot learning) from a widely acknowledged benchmark, LongBench \cite{bai2023longbench}, and measure them against models reported in the LongBench leaderboard. However, the maximum average document length of test sets in LongBench is shorter than 20k words, which is not very long for modern long-context models. To better understand HMT's long-context processing ability under various context scenarios, we further study HMT on crafted and controllable dataset samples. For crafted datasets, we derive from existing datasets to form long inputs. For general language tasks, models are tested for next token generation tasks with Wikitext-103 \cite{merity2016pointer} (2-3k words per sample) and PG-19 \cite{rae2019compressive} datasets (69k words per sample on average). Samples will be concatenated or split into chunks to form longer samples and investigate the relationships between input length and the effectiveness of the model. For question-answering tasks, we chose PubMedQA \cite{jin2019pubmedqa}, which is a biomedical question-answering dataset with corresponding contexts. We artifact the dataset to assess HMT with multi-context inputs, described in Appendix \ref{sec:pubmedqa}. 

\begin{figure*}[ht]
    \centering
    \includegraphics[width=0.9\textwidth]{ 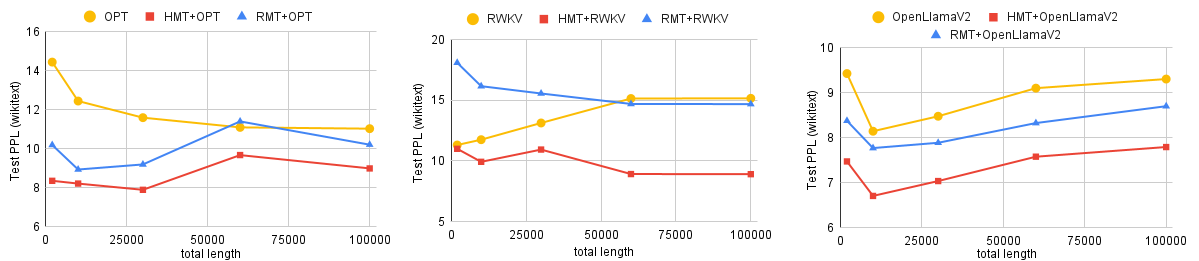}
    \caption{Test Perplexity of HMT, RMT, and three baseline models (OPT 2.7B, RWKV 3B, OpenLlamaV2 3B) with the Wikitext-103 dataset. HMT outperforms RMT by 13.0\% for OPT and 10.8\% for OpenLlamaV2. For RWKV, HMT can even boost the effectiveness by 16.5\%, while RMT worsens the effectiveness.}
    \label{fig:compare1}
\end{figure*}

\begin{figure*}[ht]
    \centering
    \includegraphics[width=0.9\textwidth]{ 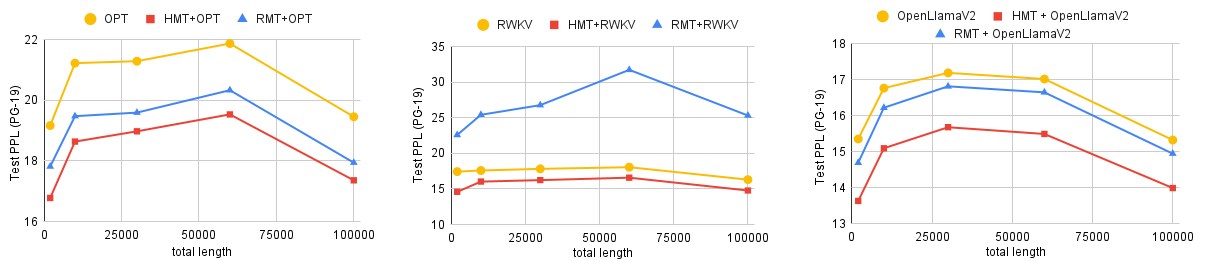}
    \caption{Test Perplexity of HMT, RMT, and three baseline models (OPT 2.7B, RWKV 3B, OpenLlamaV2 3B), evaluated over the PG-19 dataset. HMT outperforms RMT by 3.98\% for OPT and 6.85\% for OpenLlamaV2. For RWKV, HMT can improve the effectiveness by 9.96\%.}
    \label{fig:compare2}
\end{figure*}

\section{Results and Key Observations} \label{sec:exp_res}

In this section, we illustrate the main result of HMT. More ablation studies are in Appendix \ref{sec:ablation} and \ref{sec:mem_behavior}.

\subsection{Impacts on Backbone Models} \label{sec:res}

By introducing an additional 0.5\% $\sim$ 1.3\% (1.77M $\sim$ 33.5M) of parameters, HMT can enhance models with a variety of architectures to improve generation quality when processing long context inputs. We demonstrate this feature with general language modeling and question-answering tasks.

\begin{figure}[ht]
    \centering
    \includegraphics[width=0.65\columnwidth]{ 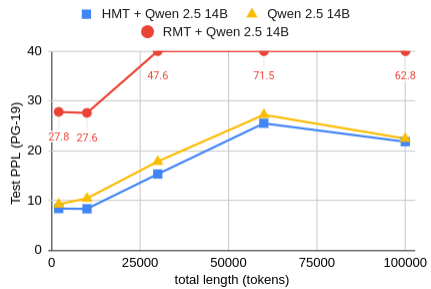}
    \caption{Test Perplexity of HMT, RMT, and baseline model for Qwen 2.5 14B on PG-19 dataset. HMT boosts the effectiveness of the baseline model by 10.0\%, while RMT worsens its effectiveness.}
    \label{fig:qwen14b}
\end{figure}

\textbf{HMT consistently improves the backbone models in general language modeling tasks when processing long inputs.} Figures \ref{fig:compare1} and \ref{fig:compare2} compare the perplexity of OPT 2.7B, RWKV 3B, and OpenLlamaV2 3B models with and without HMT on the Wikitext-103 and PG-19 datasets. Over input spanning from 2k $\sim$ 100k tokens, HMT consistently raises the generation quality of all these models. Moreover, Table \ref{tab:scale-ppl} presents how improvements are achieved by HMT scales with the model size for same-family models. To further strengthen our argument that HMT can benefit larger models, we evaluate HMT with Qwen 2.5 14B utilizing 4 A100-80GB GPUs for training. As depicted in Figure \ref{fig:qwen14b}, HMT can still increase the effectiveness of the backbone model on PG-19.

Notice that the improvement is not necessarily contributed solely by the additional parameters. Having more parameters does not always lead to higher performance. For example, HMT boosts OPT 2.7B to realize a lower perplexity than OpenLlama 3B with 20.7\% fewer parameters, while OPT 2.7B performs worse without HMT. Section \ref{sec:long-context} describes more examples of HMT achieving superior generation quality with smaller models.

\begin{table}
\caption{Scalability of HMT. Average PPL is computed by taking the average PPL for samples in each sequence length in the experiment.}
\label{tab:scale-ppl}
\begin{center}
\begin{small}
\begin{sc}
\resizebox{0.75\columnwidth}{!}{
\begin{tabular}{ccc}
\toprule
Model & Avg Test PPL (Wikitext) ($\downarrow$) \\
\midrule
OPT 350M & 15.11\\
HMT + OPT 350M & 14.28 (-5.8\%)\\
\midrule
OPT 2.7B & 12.12\\
HMT + OPT 2.7B & 8.61 (-28.9\%) \\
\bottomrule
\toprule
RWKV 430M & 19.33 \\
HMT + RWKV 430M & 16.10 (-16.6\%)\\
\midrule
RWKV 3B & 13.30 \\
HMT + RWKV 3B & 9.93 (-25.3\%)\\
\bottomrule
\end{tabular}
}
\end{sc}
\end{small}
\end{center}
\end{table}

\begin{figure}[ht]
    \centering
    \includegraphics[width=0.65\columnwidth]{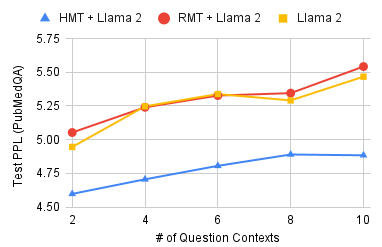}
    \caption{Long answer quality of RMT and HMT applied on Llama-2 7B, evaluated over PubMedQA dataset. HMT is 8.98\% more effective than RMT.}
    \label{fig:qa1}
\end{figure}

\begin{figure}[ht]
    \centering
    \includegraphics[width=0.65\columnwidth]{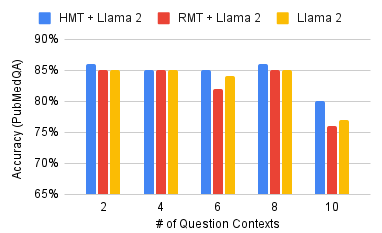}
    \caption{Short response accuracy of RMT and HMT applied on Llama-2 7B, evaluated over PubMedQA dataset. HMT is 1.8\% more accurate than RMT.}
    \label{fig:qa2}
\end{figure}

\textbf{HMT enhances long-answer contextual reasoning and short-answer prediction ability in question-answering tasks.} One of the use cases of HMT is handling question-answering tasks that involve multiple contexts. Thus, we select the PubMedQA dataset and derive long-context QA samples with controllable context counts to evaluate the effectiveness of HMT. Two metrics are employed: for long answers, we compute the PPL to assess the contextual reasoning of HMT; for short answers, we measure the response accuracy. As seen in Figures \ref{fig:qa1} and \ref{fig:qa2}, for samples with 2 to 10 contexts, HMT increases the effectiveness in PPL by 9.48\% for long answers. For short answer tasks, HMT is 1.0\% more accurate than the backbone model and exhibits significant advantages when samples have more contexts. In sum, HMT increases both the correctness and reasoning ability of models in long-context QA tasks.

\subsection{Comparison to Long Context Models} \label{sec:long-context}

Combined with small and short-context models, HMT can be more effective than large models trained on long-context inputs. Table \ref{tab:longbench} displays metric results of HMT-augmented models on subsets of LongBench \cite{bai2023longbench} and compares them with large models specialized for long contexts. The subsets contain various generation tasks, including single/multi-document QA, summarization, and few-shot learning. With a significantly lower inference memory requirement, HMT applied to small models can attain comparable or better metrics compared to large models, indicating a significant resource advantage. Specifically, we observe that HMT with small models performs well in generating short responses for long and multi-context inputs, thanks to its context-filtering ability. However, it exhibits comparable or weaker performance in generating long responses, as small models have shorter token generation limits compared to large models.

Moreover, applying HMT to long-context models can further improve their effectiveness and reduce inference memory consumption. For example, the AMD MI210 GPU cannot handle inferencing 30k token inputs with the Yi-6B-200K model due to memory constraints. Applying a sliding window strategy with a 5.2K-token window (Yi-6B-SW-5.2K), the model consumes 44.8 GB VRAM. On the contrary, HMT + Yi-6B-200K requires only 33.9 GB VRAM to process 30k tokens with a small segment length (512 tokens), with a 2\% effectiveness improvement. Table \ref{tab:long_ppl} presents the effectiveness of long-range models on Wikitext-103 compared with several HMT-augmented models, including Mamba and Mistral models. 

\begin{table*}
\caption{Metric results of HMT-augmented small models and large models trained on longer contexts. Models with HMT can process infinitely long context, but only keep a fixed length of KV cache (the value in the parenthesis). We evaluate on subsets of LongBench, including QMSum (QMS), MuSiQue (MSQ), Qasper (QASP), NarrativeQA (NQA), MultiFieldQA-en (MFQA-en), GovReport (GR), TriviaQA (TQA), HotpotQA (HQA), 2WikiMQA (2WMQA), and MultiNews (MN). Mem Req indicates the minimum inference memory required (to store parameters and KV cache). Actual inference may require a larger VRAM.}
\label{tab:longbench}
\begin{center}
\begin{small}
\begin{sc}
\resizebox{\textwidth}{!}{
\begin{tabular}{ccccccccccccc}
\toprule
Model & Context Window & QMS & MSQ & QASP & NQA & MFQA-en & GR & TQA & HQA & 2WMQA & MN & Mem Req (GB $\downarrow$) \\
\midrule
GPT-3.5 Turbo & 16384 & 23.4 & 26.9 & \textbf{43.3} & 23.6 & 52.3 & 29.5 & 91.4 & 51.6 & 37.7 & 26.7 & 46.7\\
Llama 2 7B Chat & 4096 & 20.8 & 9.4 & 19.2 & 18.7 & 36.8 & 27.3 & 77.8 & 25.4 & 32.8 & 25.8 & 15.1\\
LongChat V1.5 7B & 32768 & 22.7 & 9.7 & 27.7 & 16.9 & 41.4 & 30.8 & 82.3 & 31.5 & 20.6 & 26.4 & 22.6\\
XGen 7B & 8192 & 20.5 & 10.3 & 18.1 & 18.0 & 37.7 & 27.3 & 77.8 & 29.7 & 21.1& 26.2& 16.2 \\
InternLM 7B & 8192 & 15.9 & 9.0 & 16.7 & 12.1 & 23.4 & 9.7 & 77.8 & 28.7 & 21.1& 22.8& 16.2 \\
ChatGLM2 6B & 32768 & \textbf{24.0} & 21.9 & 31.5 & 21.1 & 46.2 & 32.4 & 78.7 & 45.1 & 34.0& 26.5& 19.5 \\
Vicuna V1.5 7B & 16384 & 22.8 & 9.8 & 26.1 & 19.4 & 38.5 & 27.9 & 86.2 & 25.3 & 20.8& 27.2 & 18.3 \\
ChatGLM3 6B & 32768 & 23.9 & 40.4 & \textbf{43.3 }& \textbf{26.0} & 51.7 & \textbf{36.8} & 87.1 & \textbf{54.4} & \textbf{44.9}& \textbf{27.9}& 19.5\\
\midrule
HMT + OPT 350M & $\infty$ (1024) & 22.2 & 15.6 & 32.3 & 17.2 & \textbf{62.7} & 13.8 & 87.2 &25.3 & 29.2 & 10.5 & 0.75\\
HMT + OpenLlamaV2 3B & $\infty$ (512) &21.4 & \textbf{42.3} & 35.7 & 21.6 & 58.1 & 15.8 & \textbf{93.0}& 29.9 & 27.7 & 11.0 & 6.1 \\
HMT + SmolLM 135M & $\infty$ (1024) & 19.4 & 14.1 & 30.2 & 19.6 & 48.1 & 15.7 & 81.3 & 27.6 & 23.3 & 9.5 & \textbf{0.4}\\
\bottomrule
Benchmark Average Len & (words) & 10614 & 11214 & 3619 & 18409& 4559& 8734& 8209& 9151& 4887& 2113&\\
\bottomrule
\end{tabular}
}
\end{sc}
\end{small}
\end{center}
\end{table*}

\begin{table}
\caption{Quality of long context models and HMT with various backbone models. The input size is 30k tokens and the dataset is Wikitext-103.}
\label{tab:long_ppl}
\begin{center}
\begin{small}
\begin{sc}
\resizebox{0.9\columnwidth}{!}{
\begin{tabular}{ccc}
\toprule
Model & Max Context & Test PPL (Wikitext) \\
\midrule
RWKV 3B & $\infty$ & 13.13 \\
Mamba 370M & $\infty$ & 87.08 \\
Yi-6B-200K & 200K & OOM \tablefootnote{Although this model is trained with 200K-token samples, it cannot be run on MI210 due to memory constraints.} \\
Yi-6B-SW-5.2K & 200K & 6.89 \\
Mistral-7B & 32K & 5.47 \\
\midrule
HMT + OPT 350M & $\infty$ (1024) & 13.67 \\
HMT + OpenLlamaV2 3B & $\infty$ (512) & 7.04 \\
HMT + RWKV 3B & $\infty$ (256)& 10.94 \\
HMT + Mamba 370M & $\infty$ (256)& 16.71 \\
HMT + Yi-6B-200K & $\infty$ (512)& 6.75 \\
HMT + Mistral-7B & $\infty$ (512)& 5.12 \\
\bottomrule
\end{tabular}
}
\end{sc}
\end{small}
\end{center}
\end{table}

\subsection{Comparison to Memory-augmented and Hierarchical Methods}

One popular memory-augmented model is the recurrent memory transformer \cite{bulatov2022recurrent} (RMT). Our assessment indicates that HMT is generally better at both language modeling and question-answering tasks than RMT, illustrated in Figures \ref{fig:compare1}, \ref{fig:compare2}, \ref{fig:qa1}, and \ref{fig:qa2}. The improvement gap is especially significant for recurrent models such as RWKV. HMT can further increase the effectiveness of RWKV while RMT will degrade the performance for both datasets, as demonstrated in Figure \ref{fig:compare2}. Since RWKV has already compressed past tokens and passed hidden states along the sequence, applying RMT to RWKV re-weights past information compressed in states periodically. This was originally done by the time-mixing module of RWKV. Therefore, the advantage of memory augmentation is limited. Due to the gradient vanishing issue, the model is harder to train with RMT, leading to inferior performance. However, we believe that the memory retrieval mechanism in HMT helps RWKV to select previous hidden states with the most relevance, boosting its effectiveness. Another advantage of HMT over RMT is its scalability with large models: while RMT applied to Qwen 2.5 14B results in reduced effectiveness compared to direct inference with the backbone model, HMT continues to enhance effectiveness, as illustrated in Figure \ref{fig:qwen14b}.

Furthermore, compared with other memory-augmented models, HMT is not only easy to use but also has higher generation quality. Table \ref{tab:memtrm} picks three memory-augmented methods (Memorizing Transformer \cite{wu2022memorizing}, LongMem \cite{wang2024augmenting}, and CCM-concat \cite{kim2023compressed}) and compares them with HMT with the same or similar-sized backbone models. We choose the datasets used by the original works for fair comparisons. Memorizing transformer and LongMem require modifying the core architecture of the base model. Future models cannot easily adopt such modifications. Overall, HMT outperforms these methods. We also list the inference memory overhead complexity for each model, where $L$ is the total context length, $l_i$ is the inference segment length, $l_m$ is the memory size ($L > l_m > l_i$), and $t$ is the number of memory embeddings concatenated for CCM-concat. HMT has the lowest memory complexity over all previous methods.

\begin{figure}[ht]
    \centering
    \includegraphics[width=0.75\columnwidth]{ 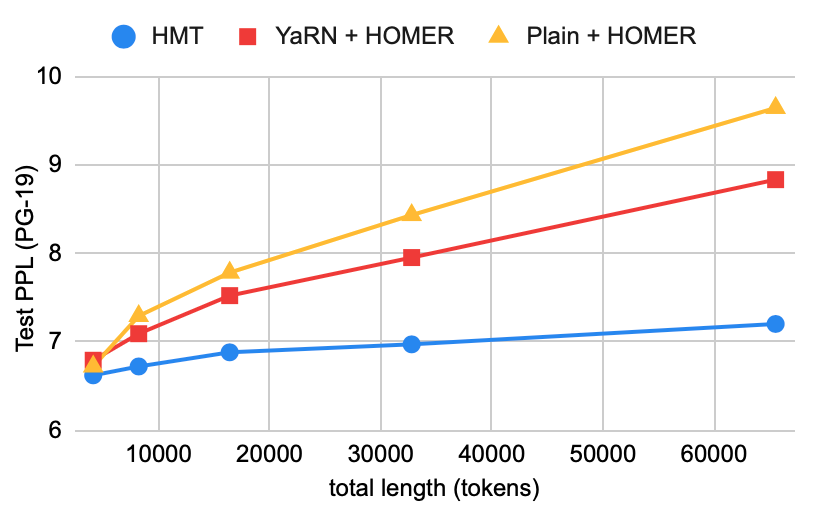}
    \caption{Comparison between HMT and HOMER without context extension and with YaRN, all applying on Llama 2 7B. On average, HMT is 9.9\% more effective than HOMER with YaRN on PG-19.}
    \label{fig:homer}
\end{figure}

Lastly, we compare HMT with HOMER \cite{song2024hierarchical}, a method that hierarchically compresses inputs to reduce their length for inference. In terms of memory complexity, HOMER requires $O(\log(L))$ memory to store the reduction tree, leading to increased peak memory utilization as input length grows. In contrast, HMT maintains a constant peak memory complexity regardless of input length. Regarding effectiveness, HMT achieves 9.9\% lower perplexity on PG-19 compared to HOMER with YaRN \cite{peng2023yarn} for context extension. As shown in Figure \ref{fig:homer}, the benefits of HMT become more substantial as input length increases, highlighting its superior scalability with longer inputs.

\begin{table}
\caption{Comparison between HMT with previous memory-augmented methods (Memorizing Transformer, LongMem, and CCM-concat).}
\label{tab:memtrm}
\begin{center}
\begin{small}
\begin{sc}
\resizebox{\columnwidth}{!}{
\begin{tabular}{ccc}
\toprule
Model & Test PPL (Wikitext, 30k token) & Mem Overhead \\
\midrule
MemTRM & 31.51 & $O(L)$\\
HMT + OPT 350M & \textbf{13.67} & $O(l_i)$\\
\bottomrule
\toprule
Model & Test PPL (ArXiv, variable) & Mem Overhead\\
\midrule
LongMem & 10.08 & $O(l_m)$\\
HMT + Qwen1.5-0.5B & \textbf{9.02} & $O(l_i)$\\
\bottomrule
\toprule
Model & Test PPL (PG-19, 60k token) & Mem Overhead \\
\midrule
CCM-concat & 7.41 & $O(t+l_i)$\\
HMT + Llama 2 7B & \textbf{7.40} & $O(l_i)$\\
\bottomrule
\end{tabular}
}
\end{sc}
\end{small}
\end{center}
\vskip -0.1in
\end{table}



\section{Conclusion} \label{sec:conclusion}

We present HMT, a framework to augment models' long-range language processing ability with context switching. Inspired by the brain's memory hierarchy, HMT imitates human memorization behavior by deploying hierarchical memory and the memory retrieval mechanism. HMT consistently improves the generation quality of the backbone models. Compared with other long-context LLMs and memory-augmented models, HMT achieves higher generation quality with lower memory requirements. Our model provides LLM accessibility to resource-constrained applications and represents a step forward to lifelong language tasks.

\section{Limitations and Ongoing Works}

\begin{itemize}[wide=0pt]
    \item Currently, HMT will save $N$ memory embeddings for memory search, which is a cross-attention layer. When $N$ is small (e.g., $N = 300$), which is already sufficient for 100k token samples, the overhead is negligible. However, when $N$ grows and the memory embeddings are stored in different physical memory hierarchies, the overhead can be significant. An intelligent memory prefetching mechanism can potentially alleviate the latency overhead, which we leave as future work.
    \item Due to the large computational graph of models when training with BPTT, tuning the extra parameters introduced by HMT can be memory-consuming, impeding experiments on larger-scale models. A more efficient way to extend BPTT depth without memory overhead is a future research direction.
    \item Although HMT employs only one level of long-term memory, one may use multiple levels of long-term memory to improve information access efficiency. Similar techniques have been used for multilevel optimization in VLSI physical design \cite{cong2013multilevel, chan2005multilevel}.
\end{itemize}

\section{Ethical Statements} \label{sec:impact}
The capability of memorizing information by HMT offers convenience to people's daily lives, while also raising concerns about privacy leakage through conversation with language model agents. Nevertheless, with further efforts to deploy it on edge devices without network connections, this issue can be resolved.

\section*{Acknowledgement}
This research is partially supported by the PRISM (000705769) center under the JUMP 2.0 program by DARPA/SRC and NSF SEED funding. It is also supported by CDSC industrial partners (\url{https://cdsc.ucla.edu/partners}) and the AMD HACC Program.

\bibliography{custom}

\begin{thebibliography}{57}
\providecommand{\natexlab}[1]{#1}

\bibitem[{Agerri et~al.(2015)Agerri, Artola, Beloki, Rigau, and Soroa}]{agerri2015big}
Rodrigo Agerri, Xabier Artola, Zuhaitz Beloki, German Rigau, and Aitor Soroa. 2015.
\newblock Big data for natural language processing: A streaming approach.
\newblock \emph{Knowledge-Based Systems}, 79:36--42.

\bibitem[{Allal et~al.(2024)Allal, Lozhkov, Bakouch, von Werra, and Wolf}]{allal2024SmolLM}
Loubna~Ben Allal, Anton Lozhkov, Elie Bakouch, Leandro von Werra, and Thomas Wolf. 2024.
\newblock Smollm - blazingly fast and remarkably powerful.

\bibitem[{Bahdanau et~al.(2014)Bahdanau, Cho, and Bengio}]{bahdanau2014neural}
Dzmitry Bahdanau, Kyunghyun Cho, and Yoshua Bengio. 2014.
\newblock Neural machine translation by jointly learning to align and translate.
\newblock \emph{arXiv preprint arXiv:1409.0473}.

\bibitem[{Bai et~al.(2023{\natexlab{a}})Bai, Bai, Chu, Cui, Dang, Deng, Fan, Ge, Han, Huang et~al.}]{bai2023qwen}
Jinze Bai, Shuai Bai, Yunfei Chu, Zeyu Cui, Kai Dang, Xiaodong Deng, Yang Fan, Wenbin Ge, Yu~Han, Fei Huang, et~al. 2023{\natexlab{a}}.
\newblock Qwen technical report.
\newblock \emph{arXiv preprint arXiv:2309.16609}.

\bibitem[{Bai et~al.(2023{\natexlab{b}})Bai, Lv, Zhang, Lyu, Tang, Huang, Du, Liu, Zeng, Hou et~al.}]{bai2023longbench}
Yushi Bai, Xin Lv, Jiajie Zhang, Hongchang Lyu, Jiankai Tang, Zhidian Huang, Zhengxiao Du, Xiao Liu, Aohan Zeng, Lei Hou, et~al. 2023{\natexlab{b}}.
\newblock Longbench: A bilingual, multitask benchmark for long context understanding.
\newblock \emph{arXiv preprint arXiv:2308.14508}.

\bibitem[{Beltagy et~al.(2020)Beltagy, Peters, and Cohan}]{beltagy2020longformer}
Iz~Beltagy, Matthew~E Peters, and Arman Cohan. 2020.
\newblock Longformer: The long-document transformer.
\newblock \emph{arXiv preprint arXiv:2004.05150}.

\bibitem[{Bertsch et~al.(2023)Bertsch, Alon, Neubig, and Gormley}]{bertsch2023unlimiformer}
Amanda Bertsch, Uri Alon, Graham Neubig, and Matthew~R Gormley. 2023.
\newblock Unlimiformer: Long-range transformers with unlimited length input.
\newblock \emph{arXiv preprint arXiv:2305.01625}.

\bibitem[{Bulatov et~al.(2022)Bulatov, Kuratov, and Burtsev}]{bulatov2022recurrent}
Aydar Bulatov, Yury Kuratov, and Mikhail Burtsev. 2022.
\newblock Recurrent memory transformer.
\newblock \emph{Advances in Neural Information Processing Systems}, 35:11079--11091.

\bibitem[{Burgin(2011)}]{burgin2011epistemic}
Mark Burgin. 2011.
\newblock Epistemic information in stratified m-spaces.
\newblock \emph{Information}, 2(4):697--726.

\bibitem[{Chan et~al.(2005)Chan, Cong, and Sze}]{chan2005multilevel}
Tony Chan, Jason Cong, and Kenton Sze. 2005.
\newblock Multilevel generalized force-directed method for circuit placement.
\newblock In \emph{Proceedings of the 2005 international symposium on physical design}, pages 185--192.

\bibitem[{Chang et~al.(2015)Chang, Martini, and Culurciello}]{chang2015recurrent}
Andre Xian~Ming Chang, Berin Martini, and Eugenio Culurciello. 2015.
\newblock Recurrent neural networks hardware implementation on fpga.
\newblock \emph{arXiv preprint arXiv:1511.05552}.

\bibitem[{Chevalier et~al.(2023)Chevalier, Wettig, Ajith, and Chen}]{chevalier2023adapting}
Alexis Chevalier, Alexander Wettig, Anirudh Ajith, and Danqi Chen. 2023.
\newblock Adapting language models to compress contexts.
\newblock \emph{arXiv preprint arXiv:2305.14788}.

\bibitem[{Chung et~al.(2014)Chung, Gulcehre, Cho, and Bengio}]{chung2014empirical}
Junyoung Chung, Caglar Gulcehre, KyungHyun Cho, and Yoshua Bengio. 2014.
\newblock Empirical evaluation of gated recurrent neural networks on sequence modeling.
\newblock \emph{arXiv preprint arXiv:1412.3555}.

\bibitem[{Computer(2023)}]{together2023redpajama}
Together Computer. 2023.
\newblock \href {https://github.com/togethercomputer/RedPajama-Data} {Redpajama: an open dataset for training large language models}.

\bibitem[{Cong and Shinnerl(2013)}]{cong2013multilevel}
Jingsheng~Jason Cong and Joseph~R Shinnerl. 2013.
\newblock \emph{Multilevel optimization in VLSICAD}, volume~14.
\newblock Springer Science \& Business Media.

\bibitem[{Dai et~al.(2022)Dai, Lang, Zheng, Huang, Si, and Li}]{dai2022lifelong}
Yi~Dai, Hao Lang, Yinhe Zheng, Fei Huang, Luo Si, and Yongbin Li. 2022.
\newblock Lifelong learning for question answering with hierarchical prompts.
\newblock \emph{arXiv preprint arXiv:2208.14602}.

\bibitem[{Dai et~al.(2019)Dai, Yang, Yang, Carbonell, Le, and Salakhutdinov}]{dai2019transformer}
Zihang Dai, Zhilin Yang, Yiming Yang, Jaime Carbonell, Quoc~V Le, and Ruslan Salakhutdinov. 2019.
\newblock Transformer-xl: Attentive language models beyond a fixed-length context.
\newblock \emph{arXiv preprint arXiv:1901.02860}.

\bibitem[{Dao et~al.(2022)Dao, Fu, Ermon, Rudra, and R{\'e}}]{dao2022flashattention}
Tri Dao, Dan Fu, Stefano Ermon, Atri Rudra, and Christopher R{\'e}. 2022.
\newblock Flashattention: Fast and memory-efficient exact attention with io-awareness.
\newblock \emph{Advances in Neural Information Processing Systems}, 35:16344--16359.

\bibitem[{Dong et~al.(2019)Dong, Yang, Wang, Wei, Liu, Wang, Gao, Zhou, and Hon}]{dong2019unified}
Li~Dong, Nan Yang, Wenhui Wang, Furu Wei, Xiaodong Liu, Yu~Wang, Jianfeng Gao, Ming Zhou, and Hsiao-Wuen Hon. 2019.
\newblock Unified language model pre-training for natural language understanding and generation.
\newblock \emph{Advances in neural information processing systems}, 32.

\bibitem[{Dosovitskiy et~al.(2020)Dosovitskiy, Beyer, Kolesnikov, Weissenborn, Zhai, Unterthiner, Dehghani, Minderer, Heigold, Gelly et~al.}]{dosovitskiy2020image}
Alexey Dosovitskiy, Lucas Beyer, Alexander Kolesnikov, Dirk Weissenborn, Xiaohua Zhai, Thomas Unterthiner, Mostafa Dehghani, Matthias Minderer, Georg Heigold, Sylvain Gelly, et~al. 2020.
\newblock An image is worth 16x16 words: Transformers for image recognition at scale.
\newblock \emph{arXiv preprint arXiv:2010.11929}.

\bibitem[{Dubey et~al.(2024)Dubey, Jauhri, Pandey, Kadian, Al-Dahle, Letman, Mathur, Schelten, Yang, Fan et~al.}]{dubey2024llama}
Abhimanyu Dubey, Abhinav Jauhri, Abhinav Pandey, Abhishek Kadian, Ahmad Al-Dahle, Aiesha Letman, Akhil Mathur, Alan Schelten, Amy Yang, Angela Fan, et~al. 2024.
\newblock The llama 3 herd of models.
\newblock \emph{arXiv preprint arXiv:2407.21783}.

\bibitem[{Gao et~al.(2020)Gao, Biderman, Black, Golding, Hoppe, Foster, Phang, He, Thite, Nabeshima et~al.}]{gao2020pile}
Leo Gao, Stella Biderman, Sid Black, Laurence Golding, Travis Hoppe, Charles Foster, Jason Phang, Horace He, Anish Thite, Noa Nabeshima, et~al. 2020.
\newblock The pile: An 800gb dataset of diverse text for language modeling.
\newblock \emph{arXiv preprint arXiv:2101.00027}.

\bibitem[{Geng and Liu(2023)}]{openlm2023openllama}
Xinyang Geng and Hao Liu. 2023.
\newblock \href {https://github.com/openlm-research/open_llama} {Openllama: An open reproduction of llama}.

\bibitem[{Gu and Dao(2023)}]{gu2023mamba}
Albert Gu and Tri Dao. 2023.
\newblock Mamba: Linear-time sequence modeling with selective state spaces.
\newblock \emph{arXiv preprint arXiv:2312.00752}.

\bibitem[{Hochreiter and Schmidhuber(1997)}]{hochreiter1997long}
Sepp Hochreiter and J{\"u}rgen Schmidhuber. 1997.
\newblock Long short-term memory.
\newblock \emph{Neural computation}, 9(8):1735--1780.

\bibitem[{Hu et~al.(2021)Hu, Shen, Wallis, Allen-Zhu, Li, Wang, Wang, and Chen}]{hu2021lora}
Edward~J Hu, Yelong Shen, Phillip Wallis, Zeyuan Allen-Zhu, Yuanzhi Li, Shean Wang, Lu~Wang, and Weizhu Chen. 2021.
\newblock Lora: Low-rank adaptation of large language models.
\newblock \emph{arXiv preprint arXiv:2106.09685}.

\bibitem[{Jiang et~al.(2023)Jiang, Sablayrolles, Mensch, Bamford, Chaplot, Casas, Bressand, Lengyel, Lample, Saulnier et~al.}]{jiang2023mistral}
Albert~Q Jiang, Alexandre Sablayrolles, Arthur Mensch, Chris Bamford, Devendra~Singh Chaplot, Diego de~las Casas, Florian Bressand, Gianna Lengyel, Guillaume Lample, Lucile Saulnier, et~al. 2023.
\newblock Mistral 7b.
\newblock \emph{arXiv preprint arXiv:2310.06825}.

\bibitem[{Jin et~al.(2019)Jin, Dhingra, Liu, Cohen, and Lu}]{jin2019pubmedqa}
Qiao Jin, Bhuwan Dhingra, Zhengping Liu, William~W Cohen, and Xinghua Lu. 2019.
\newblock Pubmedqa: A dataset for biomedical research question answering.
\newblock \emph{arXiv preprint arXiv:1909.06146}.

\bibitem[{Kim et~al.(2023)Kim, Yeom, Yun, and Song}]{kim2023compressed}
Jang-Hyun Kim, Junyoung Yeom, Sangdoo Yun, and Hyun~Oh Song. 2023.
\newblock Compressed context memory for online language model interaction.
\newblock \emph{arXiv preprint arXiv:2312.03414}.

\bibitem[{Kitaev et~al.(2020)Kitaev, Kaiser, and Levskaya}]{kitaev2020reformer}
Nikita Kitaev, {\L}ukasz Kaiser, and Anselm Levskaya. 2020.
\newblock Reformer: The efficient transformer.
\newblock \emph{arXiv preprint arXiv:2001.04451}.

\bibitem[{Kovaleva et~al.(2019)Kovaleva, Romanov, Rogers, and Rumshisky}]{kovaleva2019revealing}
Olga Kovaleva, Alexey Romanov, Anna Rogers, and Anna Rumshisky. 2019.
\newblock Revealing the dark secrets of bert.
\newblock \emph{arXiv preprint arXiv:1908.08593}.

\bibitem[{Lieber et~al.(2024)Lieber, Lenz, Bata, Cohen, Osin, Dalmedigos, Safahi, Meirom, Belinkov, Shalev-Shwartz et~al.}]{lieber2024jamba}
Opher Lieber, Barak Lenz, Hofit Bata, Gal Cohen, Jhonathan Osin, Itay Dalmedigos, Erez Safahi, Shaked Meirom, Yonatan Belinkov, Shai Shalev-Shwartz, et~al. 2024.
\newblock Jamba: A hybrid transformer-mamba language model.
\newblock \emph{arXiv preprint arXiv:2403.19887}.

\bibitem[{Liu et~al.(2023)Liu, Yuan, Fu, Jiang, Hayashi, and Neubig}]{liu2023pre}
Pengfei Liu, Weizhe Yuan, Jinlan Fu, Zhengbao Jiang, Hiroaki Hayashi, and Graham Neubig. 2023.
\newblock Pre-train, prompt, and predict: A systematic survey of prompting methods in natural language processing.
\newblock \emph{ACM Computing Surveys}, 55(9):1--35.

\bibitem[{Merity et~al.(2016)Merity, Xiong, Bradbury, and Socher}]{merity2016pointer}
Stephen Merity, Caiming Xiong, James Bradbury, and Richard Socher. 2016.
\newblock Pointer sentinel mixture models.
\newblock \emph{arXiv preprint arXiv:1609.07843}.

\bibitem[{Modarressi et~al.(2023)Modarressi, Imani, Fayyaz, and Sch{\"u}tze}]{modarressi2023ret}
Ali Modarressi, Ayyoob Imani, Mohsen Fayyaz, and Hinrich Sch{\"u}tze. 2023.
\newblock Ret-llm: Towards a general read-write memory for large language models.
\newblock \emph{arXiv preprint arXiv:2305.14322}.

\bibitem[{Moro et~al.(2023)Moro, Ragazzi, Valgimigli, Frisoni, Sartori, and Marfia}]{moro2023efficient}
Gianluca Moro, Luca Ragazzi, Lorenzo Valgimigli, Giacomo Frisoni, Claudio Sartori, and Gustavo Marfia. 2023.
\newblock Efficient memory-enhanced transformer for long-document summarization in low-resource regimes.
\newblock \emph{Sensors}, 23(7):3542.

\bibitem[{Mozer(2013)}]{mozer2013focused}
Michael~C Mozer. 2013.
\newblock A focused backpropagation algorithm for temporal pattern recognition.
\newblock In \emph{Backpropagation}, pages 137--169. Psychology Press.

\bibitem[{Pascanu et~al.(2013)Pascanu, Mikolov, and Bengio}]{pascanu2013difficulty}
Razvan Pascanu, Tomas Mikolov, and Yoshua Bengio. 2013.
\newblock On the difficulty of training recurrent neural networks.
\newblock In \emph{International conference on machine learning}, pages 1310--1318. Pmlr.

\bibitem[{Peng et~al.(2023{\natexlab{a}})Peng, Alcaide, Anthony, Albalak, Arcadinho, Cao, Cheng, Chung, Grella, GV et~al.}]{peng2023rwkv}
Bo~Peng, Eric Alcaide, Quentin Anthony, Alon Albalak, Samuel Arcadinho, Huanqi Cao, Xin Cheng, Michael Chung, Matteo Grella, Kranthi~Kiran GV, et~al. 2023{\natexlab{a}}.
\newblock Rwkv: Reinventing rnns for the transformer era.
\newblock \emph{arXiv preprint arXiv:2305.13048}.

\bibitem[{Peng et~al.(2023{\natexlab{b}})Peng, Quesnelle, Fan, and Shippole}]{peng2023yarn}
Bowen Peng, Jeffrey Quesnelle, Honglu Fan, and Enrico Shippole. 2023{\natexlab{b}}.
\newblock Yarn: Efficient context window extension of large language models.
\newblock \emph{arXiv preprint arXiv:2309.00071}.

\bibitem[{Rae et~al.(2019)Rae, Potapenko, Jayakumar, and Lillicrap}]{rae2019compressive}
Jack~W Rae, Anna Potapenko, Siddhant~M Jayakumar, and Timothy~P Lillicrap. 2019.
\newblock Compressive transformers for long-range sequence modelling.
\newblock \emph{arXiv preprint arXiv:1911.05507}.

\bibitem[{Rajbhandari et~al.(2020)Rajbhandari, Rasley, Ruwase, and He}]{rajbhandari2020zero}
Samyam Rajbhandari, Jeff Rasley, Olatunji Ruwase, and Yuxiong He. 2020.
\newblock Zero: Memory optimizations toward training trillion parameter models.
\newblock In \emph{SC20: International Conference for High Performance Computing, Networking, Storage and Analysis}, pages 1--16. IEEE.

\bibitem[{Rasley et~al.(2020)Rasley, Rajbhandari, Ruwase, and He}]{rasley2020deepspeed}
Jeff Rasley, Samyam Rajbhandari, Olatunji Ruwase, and Yuxiong He. 2020.
\newblock Deepspeed: System optimizations enable training deep learning models with over 100 billion parameters.
\newblock In \emph{Proceedings of the 26th ACM SIGKDD International Conference on Knowledge Discovery \& Data Mining}, pages 3505--3506.

\bibitem[{Shi et~al.(2023)Shi, Chen, Misra, Scales, Dohan, Chi, Sch{\"a}rli, and Zhou}]{shi2023large}
Freda Shi, Xinyun Chen, Kanishka Misra, Nathan Scales, David Dohan, Ed~H Chi, Nathanael Sch{\"a}rli, and Denny Zhou. 2023.
\newblock Large language models can be easily distracted by irrelevant context.
\newblock In \emph{International Conference on Machine Learning}, pages 31210--31227. PMLR.

\bibitem[{Song et~al.(2024)Song, Oh, Mo, Kim, Yun, Ha, and Shin}]{song2024hierarchical}
Woomin Song, Seunghyuk Oh, Sangwoo Mo, Jaehyung Kim, Sukmin Yun, Jung-Woo Ha, and Jinwoo Shin. 2024.
\newblock Hierarchical context merging: Better long context understanding for pre-trained llms.
\newblock \emph{arXiv preprint arXiv:2404.10308}.

\bibitem[{Sun et~al.(2019)Sun, Ho, and Lee}]{sun2019lamol}
Fan-Keng Sun, Cheng-Hao Ho, and Hung-Yi Lee. 2019.
\newblock Lamol: Language modeling for lifelong language learning.
\newblock \emph{arXiv preprint arXiv:1909.03329}.

\bibitem[{Team et~al.(2024)Team, Lenz, Arazi, Bergman, Manevich, Peleg, Aviram, Almagor, Fridman, Padnos et~al.}]{team2024jamba}
Jamba Team, Barak Lenz, Alan Arazi, Amir Bergman, Avshalom Manevich, Barak Peleg, Ben Aviram, Chen Almagor, Clara Fridman, Dan Padnos, et~al. 2024.
\newblock Jamba-1.5: Hybrid transformer-mamba models at scale.
\newblock \emph{arXiv preprint arXiv:2408.12570}.

\bibitem[{Touvron et~al.(2023)Touvron, Martin, Stone, Albert, Almahairi, Babaei, Bashlykov, Batra, Bhargava, Bhosale et~al.}]{touvron2023llama}
Hugo Touvron, Louis Martin, Kevin Stone, Peter Albert, Amjad Almahairi, Yasmine Babaei, Nikolay Bashlykov, Soumya Batra, Prajjwal Bhargava, Shruti Bhosale, et~al. 2023.
\newblock Llama 2: Open foundation and fine-tuned chat models.
\newblock \emph{arXiv preprint arXiv:2307.09288}.

\bibitem[{Vaswani et~al.(2017)Vaswani, Shazeer, Parmar, Uszkoreit, Jones, Gomez, Kaiser, and Polosukhin}]{vaswani2017attention}
Ashish Vaswani, Noam Shazeer, Niki Parmar, Jakob Uszkoreit, Llion Jones, Aidan~N Gomez, {\L}ukasz Kaiser, and Illia Polosukhin. 2017.
\newblock Attention is all you need.
\newblock \emph{Advances in neural information processing systems}, 30.

\bibitem[{Wang et~al.(2024)Wang, Dong, Cheng, Liu, Yan, Gao, and Wei}]{wang2024augmenting}
Weizhi Wang, Li~Dong, Hao Cheng, Xiaodong Liu, Xifeng Yan, Jianfeng Gao, and Furu Wei. 2024.
\newblock Augmenting language models with long-term memory.
\newblock \emph{Advances in Neural Information Processing Systems}, 36.

\bibitem[{Wu et~al.(2020)Wu, Lan, Qian, Gu, Geramifard, and Yu}]{wu2020memformer}
Qingyang Wu, Zhenzhong Lan, Kun Qian, Jing Gu, Alborz Geramifard, and Zhou Yu. 2020.
\newblock Memformer: A memory-augmented transformer for sequence modeling.
\newblock \emph{arXiv preprint arXiv:2010.06891}.

\bibitem[{Wu et~al.(2022)Wu, Rabe, Hutchins, and Szegedy}]{wu2022memorizing}
Yuhuai Wu, Markus~N Rabe, DeLesley Hutchins, and Christian Szegedy. 2022.
\newblock Memorizing transformers.
\newblock \emph{arXiv preprint arXiv:2203.08913}.

\bibitem[{Yang et~al.(2024)Yang, Lin, Wang, Wu, Li, Tang, Wei, Wang, Tang, Song et~al.}]{yang2024memory3}
Hongkang Yang, Zehao Lin, Wenjin Wang, Hao Wu, Zhiyu Li, Bo~Tang, Wenqiang Wei, Jinbo Wang, Zeyun Tang, Shichao Song, et~al. 2024.
\newblock Memory3: Language modeling with explicit memory.
\newblock \emph{arXiv preprint arXiv:2407.01178}.

\bibitem[{Young et~al.(2024)Young, Chen, Li, Huang, Zhang, Zhang, Li, Zhu, Chen, Chang et~al.}]{young2024yi}
Alex Young, Bei Chen, Chao Li, Chengen Huang, Ge~Zhang, Guanwei Zhang, Heng Li, Jiangcheng Zhu, Jianqun Chen, Jing Chang, et~al. 2024.
\newblock Yi: Open foundation models by 01. ai.
\newblock \emph{arXiv preprint arXiv:2403.04652}.

\bibitem[{Zhai et~al.(2021)Zhai, Talbott, Srivastava, Huang, Goh, Zhang, and Susskind}]{zhai2021attention}
Shuangfei Zhai, Walter Talbott, Nitish Srivastava, Chen Huang, Hanlin Goh, Ruixiang Zhang, and Josh Susskind. 2021.
\newblock An attention free transformer.
\newblock \emph{arXiv preprint arXiv:2105.14103}.

\bibitem[{Zhang et~al.(2021)Zhang, Gong, Shen, Li, Lv, Duan, and Chen}]{zhang2021poolingformer}
Hang Zhang, Yeyun Gong, Yelong Shen, Weisheng Li, Jiancheng Lv, Nan Duan, and Weizhu Chen. 2021.
\newblock Poolingformer: Long document modeling with pooling attention.
\newblock In \emph{International Conference on Machine Learning}, pages 12437--12446. PMLR.

\bibitem[{Zhang et~al.(2022)Zhang, Roller, Goyal, Artetxe, Chen, Chen, Dewan, Diab, Li, Lin et~al.}]{zhang2022opt}
Susan Zhang, Stephen Roller, Naman Goyal, Mikel Artetxe, Moya Chen, Shuohui Chen, Christopher Dewan, Mona Diab, Xian Li, Xi~Victoria Lin, et~al. 2022.
\newblock Opt: Open pre-trained transformer language models.
\newblock \emph{arXiv preprint arXiv:2205.01068}.

\end{thebibliography}

\appendix
\onecolumn

\section{Other Related Works}

The memory-augmented long-context transformer has been an active research topic in recent years. LongMem \cite{wang2024augmenting} chunks the long-document input into segments and caches the attention keys and values for each segment. During the inference of a segment, LongMem will select relevant key-value embedding pairs by computing the attention score between token embeddings and the cached key embeddings and fuse the top k embeddings. Memorizing Transformer \cite{wu2022memorizing} also caches the key-value embedding pairs similar to LongMem, but utilizes a kNN search to retrieve information similar to Unlimiformer. RET-LLM \cite{modarressi2023ret} employs prompt engineering to store the informative context in a database and search keywords when the context involves questions. $\text{Memory}^3$ \cite{yang2024memory3} compresses segments of tokens into ``explicit" memory blocks and stores them directly into a memory bank for retrieval. While these works can precisely retrieve contexts, they are not scalable due to the \textbf{increasing memory consumption} of storing long contexts without compression. Segment-level recurrent models, such as EMMA \cite{moro2023efficient}, Memformer \cite{wu2020memformer}, and Transformer-XL \cite{dai2019transformer}, attempt to compress memory throughout the recurrence to reduce memory consumption. EMMA composes long-term memory from multiple short-term memory by linear combination and concatenates long and short-term memory to augment the segments. Transformer-XL propagates the compressed memory states derived from the attention of current layers to the previous layers for every iteration. Memformer augments the attention with the stored memory embeddings per time step and retrieves information using the cross-attention layer of the encoder-decoder model. However, Memformer employs a forgetting network to remove irrelevant context similar to LSTM, which can potentially delete useful contexts for unseen inputs. On the other hand, HMT condenses contexts into embeddings and retrieves information precisely without requiring a forgetting network to remove information permanently. Also, some of these works, including Memorizing Transformer, Memformer, TransformerXL, and Unlimiformer, need to fundamentally change the model architecture or inject new adapters based on different base model architecture. It makes deployment and extension to future LLMs very expensive. HMT avoids this issue by having a model-independent plug-and-play framework.

\section{Comparison to LongMem} \label{sec:longmem}

Unlike HMT, LongMem \cite{wang2024augmenting} operates on the key and value caches of each layer of the model and requires caching long caches to capture distant context. To compare with LongMem, we pick the Qwen1.5-0.5B \cite{bai2023qwen} model as the backbone model and train HMT by 700 steps with 4 segments over 100 samples of the ArXiv subset of the Pile dataset \cite{gao2020pile}. The subset has 15.4K tokens on average and 60K tokens on maximum per sample, as \cite{wang2024augmenting} described. Due to the large storage consumption of the training subset of the Pile dataset, we only extract the ArXiv subset in the validation and test split. HMT is trained on the validation set and tested on the test set. Table \ref{tab:longmem} illustrates that HMT + Qwen1.5-0.5B realizes lower PPL, with a smaller parameter size and in-memory length (number of memory embeddings, which is the number of key and value embeddings cached for LongMem). This indicates that HMT is memory efficient.

\begin{table}[ht]
\caption{Effectiveness of LongMem \cite{wang2024augmenting} and HMT + Qwen1.5-0.5B models over ArXiv subset of the Pile dataset. With HMT, the Qwen1.5-0.5B model can obtain better effectiveness with fewer parameters and shorter memory, after 700 steps of update. The result for LongMem comes from the original paper. Subscription is the standard deviation.}
\label{tab:longmem}
\vskip 0.15in
\begin{center}
\begin{small}
\begin{sc}
\resizebox{0.9\columnwidth}{!}{
\begin{tabular}{ccccc}
\toprule
Model & \# Params & Segment Length & In-memory Length & Test PPL (ArXiv) \\
\midrule
LongMem & 558M  & 1k & 65k & 10.08 \\
HMT + Qwen1.5-0.5B & 463M & 1k & 300 & $9.02_{\pm 0.04}$ \\
\bottomrule
\end{tabular}
}
\end{sc}
\end{small}
\end{center}
\vskip -0.1in
\end{table}

\section{Comparison to Unlimiformer}

There are two major differences between a previous retrieval-augmented model, Unlimiformer \cite{bertsch2023unlimiformer}, and HMT in terms of the memory retrieval mechanism:
\begin{itemize}[wide=0pt]
    \item Unlimiformer retrieves the information with kNN search over the collection of encoded token segments, while HMT uses cross-attention. We believe there are several advantages of employing cross-attention: (1) Attending Top K's most similar token segments still introduces information loss. Regarding the self-attention layer, the aggregation of tokens with less similar encodings may positively contribute to the quality of the final output. On the other hand, cross-attention fuses all cached hidden embeddings, weighted by the relative similarity, which captures the whole context. (2) The output of the cross-attention is a single embedding, which has lower computational overhead compared to attending k extra tokens.

    \item Each cached memory embedding encodes the current token segment and the previous memory embedding in HMT. Therefore, HMT can capture the whole context even with a limited number of cached embeddings. Memory recall is mainly used to rescale the importance of past information. On the other hand, the Unlimiformer needs to store all encodings, which is memory-consuming.
\end{itemize}

In terms of usage, Unlimiformer targets encoder-decoder models and injects retrieval modules into the backbone model. Although the authors recently added support for decoder-only models for token generation, only the Llama model architectures \cite{touvron2023llama} can be applied and the training/evaluation procedure is not specified. This is one of the biggest challenges for Unlimiformer to adapt to future LLMs for validation and generation. On the contrary, HMT focuses on decoder-only models. Since HMT does not inject new modules in the backbone model, it is effortless to adapt to future LLMs.

\section{HMT, RMT, and Baseline Training Details and Hyperparameters} \label{sec:hyperparam}

Table \ref{tab:config} are the training configurations of the backbone models and HMT/RMT.

\begin{table*}[t]
\caption{Training and fine-tuning configurations for the backbone models (OPT 350M, Mamba 370M, OPT 2.7B, RWKV 3B, OpenLlamaV2 3B, Llama 2 7B, Yi-6B-200K, and Mistral 7B) and the modified model after applying RMT and HMT. S1 and S2 denote the first stage and the second stage of multi-stage training for HMT. 4 AMD MI210 GPUs cannot train larger models. The given learning rate is the starting learning rate and will decay by a factor of 0.9 for OPT, OpenLlamaV2, and RWKV models and 0.7 for the remaining models for every 100 steps. The batch size is 2. }
\label{tab:config}
\vskip 0.15in
\begin{center}
\begin{small}
\begin{sc}
\resizebox{\textwidth}{!}{
\begin{tabular}{lccccc}
\toprule
Model & Input Length & Segment Length & Learning Rate & Training Steps & Extra Param\\
\midrule
OPT 350M / SmolLM 135M & 2048 & 1024 & 1e-5 & 700 &\\
\qquad +RMT & 2048 & 1024 & 1e-5 & 700 &\\
\qquad +HMT (S1) & 2048 & 1024 & 1e-5 & 200 &\\
\qquad +HMT (S2) & 15360 & 1024 & 1e-5 & 500 & 1.77M (0.5\% - 1.3\%)\\
\midrule
OPT 2.7B / OpenLlamaV2 3B & 2048 & 512 & 1e-5 & 700& \\
\qquad +RMT & 1280 & 512 & 1e-5 & 700 &\\
\qquad +HMT (S1) & 1024 & 512 & 1e-5 & 200 &\\
\qquad +HMT (S2) & 4096 & 512 & 1e-5 & 500 & 13.1M (0.5\%)\\
\midrule
RWKV 3B & 1280 & 256 & 1e-5 & 700 &\\
\qquad +RMT & 1280 & 256 & 1e-4 & 700 & \\
\qquad +HMT (S1) & 512 & 256 & 1e-5 & 200 &\\
\qquad +HMT (S2) & 1280 & 256 & 1e-5 & 500 & 13.1M (0.5\%)\\
\midrule
Llama 2 7B & 2048 & 256 & 1e-4 & 700 &\\
\qquad +RMT & 1280 & 256 & 1e-4 & 700 &\\
\qquad +HMT (S1) & 512 & 256 & 1e-4 & 200 & \\
\qquad +HMT (S2) & 2048 & 256 & 1e-4 & 500 & 33.5M (0.5\%)\\
\midrule
Mamba 370M & 1536 & 256 & 1e-4 & 700 \\
\qquad +HMT (S1) & 512 & 256 & 1e-4 & 200  \\
\qquad +HMT (S2) & 1536 & 256 & 1e-4 & 500 & 4.1M (1.1\%)\\
\midrule
Yi-6B-200K / Mistral 7B & 2048 & - & 2e-4 & 700 \\
\qquad +HMT (S1) & 1024 & 512 & 2e-4 & 200 \\
\qquad +HMT (S2) & 1536 & 512 & 2e-4 & 500 & 33.5M (0.5\%)\\
\bottomrule
\end{tabular}
}
\end{sc}
\end{small}
\end{center}
\vskip -0.1in
\end{table*}

The baselines are assessed using the sliding window attention for context-constrained models to control the memory consumption for a fair comparison. Due to the limited VRAM of GPUs, we shrink the segment length for the larger backbone model. Also, increasing memory token size does not improve the effectiveness as \cite{bulatov2022recurrent} suggested. Thus, both RMT and HMT apply memory tokens with a length of 1. For RMT, we utilize the maximum BPTT unroll depth with the best effectiveness that the GPUs can handle. HMT is trained with the multi-stage training technique illustrated in Section \ref{sec:multi-stage}. The first stage (S1) is trained with 2 segments, and the second stage (S2) is trained with the maximum BPTT unroll depth that the GPUs can manage. The size of long-term memory is 300 (i.e., $N = 300$) for OPT, SmolLM, OpenLlama, Yi, and Mistral models and 400 ($N = 400$) for the rest of the models to capture sufficient contexts and we summarize half of the segment for representation extraction (i.e., $j = L/2$). We observed that the benefit of increasing $N$ is diminishing and stops at 300 for 100k token inputs, described in Appendix \ref{sec:ablation}. We select the learning rate for HMT, RMT, and the baseline to optimize the effectiveness. Furthermore, HMT preserves 32 tokens ($k = 32$) from the previous segment as the sensory memory. All models are trained with 700 steps, which is sufficient to converge the training loss. For models with HMT, we first pretrain the model with RedPajamaV2 \cite{together2023redpajama} with 700 steps, then finetune the model for the downstream tasks with 700 steps. For LongBench, the evaluation metrics are the same as the original work and we use the Huggingface evaluate package to compute Rouge-L and F1 score. For parameter-efficient training, LoRA \cite{hu2021lora} with rank 8 is applied to models with high training memory consumption (Llama 2 7B, Mamba 370M, Yi-6B-200K, and Mistral 7B). For the rest of the models, we finetune all parameters in the backbone model. All experiments are done with 3 random seeds and we take the average metrics.

\section{Ablation Study} \label{sec:ablation}

We conduct ablation studies regarding the memory retrieval mechanism to demonstrate that (1) memory retrieval is beneficial, (2) partial summarization of a segment in memory retrieval can speed up inference while maintaining similar effectiveness, and (3) caching more memory embedding can raise the effectiveness of HMT.

\begin{figure}[ht]
    \centering
    \begin{minipage}{0.45\textwidth}
        \includegraphics[width=0.9\columnwidth]{ 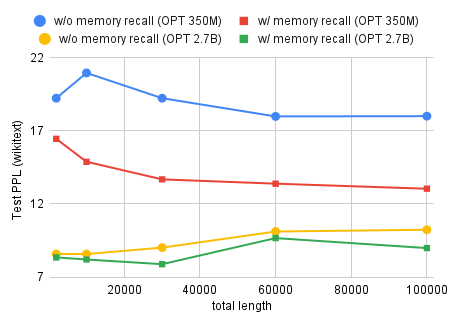}
        \caption{Effectiveness of HMT with and without the memory retrieval mechanism for OPT 350M and 2.7B as the backbone models. The inputs are extracted from the Wikitext-103 dataset with up to 100k tokens.}
        \label{fig:long-term}
    \end{minipage} \hfill
    \begin{minipage}{0.45\textwidth}
         \includegraphics[width=0.9\columnwidth]{ 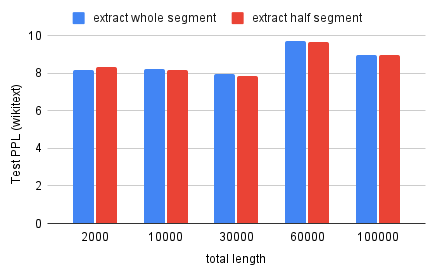}
        \caption{Effectiveness of HMT with OPT 2.7B when performing representation extraction on the whole segment for half of the segment. The impact is negligible, justifying that summarizing half of the segment is a valid method for inference acceleration.}
        \label{fig:sum-half}
    \end{minipage}
\end{figure}

\textbf{Impact of memory retrieval mechanism.} Figure \ref{fig:long-term} displays the advantages of having a memory retrieval mechanism in HMT for long context input with context switching. For any tested input length, the effectiveness of HMT with memory retrieval outperforms that without memory retrieval. Furthermore, when the memory retrieval mechanism is deployed, the effectiveness improves for the OPT 350M backbone model or tends to improve for the OPT 2.7B backbone model as the input sequence length grows, demonstrating better scalability of HMT.

\textbf{Impact of summarizing partial segment in memory retrieval.} To overlap or reduce the inference time of the previous segment with the representation extraction of the next segment, it is necessary to prefetch only the first $l$ tokens in the segment for summarization. In the experiment, we select half of the segment for representation extraction. We examine the model that extracts the whole segment and compare the effectiveness, depicted by Figure \ref{fig:sum-half}. The impact is negligible. We hypothesize that the start of a segment contains enough information about the overall topic for memory retrieval, which is intuitive as humans can capture the concepts by browsing keywords or segments instead of reading the whole paragraphs. 

\begin{figure}[ht]
    \centering
    \begin{minipage}{0.45\textwidth}
        \includegraphics[width=0.9\columnwidth]{ 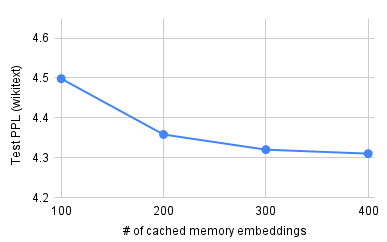}
        \caption{Relationship between number of cached memory embeddings and the effectiveness of HMT + Llama 2 7B. Each sample has 100k tokens from the Wikitext-103 dataset. As HMT stores more memory embeddings, the effectiveness is marginally better.}
        \label{fig:tradeoff}
    \end{minipage} \hfill
    \begin{minipage}{0.45\textwidth}
        \includegraphics[width=0.9\columnwidth]{ 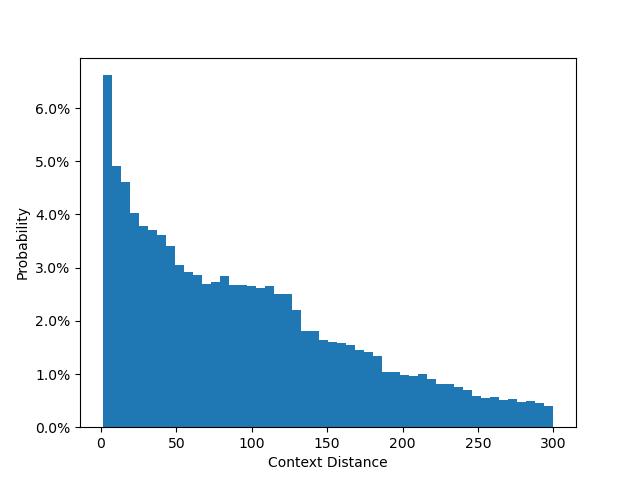}
        \caption{Histogram of context distance between the current segment and the segment corresponding to the memory embedding with the highest softmax score in the memory retrieval mechanism. The dataset evaluated is the Wikitext-103.}
        \label{fig:heatmap}
    \end{minipage}
\end{figure}

\textbf{Impact of limited cached memory embeddings.} Due to memory constraints, we only cache the most recent 300 memory embeddings for memory retrieval. Figure \ref{fig:tradeoff} depicts the relationship between the number of cached memory embeddings and the effectiveness of HMT with Llama 2 7B over the Wikitext-103 dataset with 100k-token samples. We observed that increasing the window of cached memory benefits the effectiveness, but the improvement becomes marginal. We hypothesize that HMT is more likely to recall recent memory embeddings in the Wikitext-103 dataset. Figure \ref{fig:heatmap} plots the frequency distribution of the distance between the current segment and the segment corresponding to the memory embedding with the highest softmax score in the memory retrieval mechanism. 6.5\% of the segments retrieve memory tokens within 2 segments. This signals the importance of local context. However, the long-context memory retrieval still exists. A possible explanation is that entries in Wikipedia may refer to other entries through hyperlinks and related context, and HMT discovers this long-context relationship and recalls the relevant information. 

In our experiments, we store the most recent 300 memory embeddings to balance the trade-off between retrieval effectiveness and computational efficiency. Theoretically, storing $\lceil M / L \rceil$ embeddings is sufficient to handle up to $M$ token inputs, ensuring robust performance. For longer inputs, high-quality processing is still achievable if recent contextual information is more relevant to the prompt.


\begin{figure}[ht]
    \centering
    \includegraphics[width=0.4\columnwidth]{ 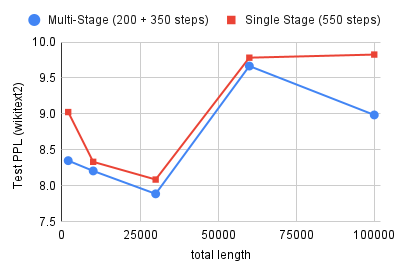}
    \caption{Training HMT + OPT 2.7B with the memory retrieval mechanism in two steps results in a better performance than using the mechanism to train HMT directly. Total training time is 902 s for multi-stage training and 1680 s for single-stage training on 4 AMD MI210 GPUs.}
    \label{fig:multi-stage}
\end{figure}

\section{Multi-stage Training} \label{sec:multi-stage}
Since HMT introduces parameters for memory retrieval, we need to train new parameters and fine-tune the parameters of the backbone model cooperatively. Training HMT involves multiple segments of tokens to learn how to encode input tokens and retrieve information properly. Therefore, we split training HMT into two stages. In the first stage, The model is trained without the memory retrieval mechanism employing BPTT with 2 segments unrolled. BPTT saves the model checkpoint locally. Then, the memory retrieval mechanism loads and extends the train model in the second stage. At this point, BPTT trains the modified model by unrolling the maximum number of segments that the GPUs can handle to maximize the effectiveness, which is 15 in our experiment. Since the architecture of HMT is complex, breaking up training into two stages is beneficial for local optimization and improves long context inference performance compared with single-stage training. Figure \ref{fig:multi-stage} exhibits the performance difference between the multi-stage and single-stage training of the OPT 2.7B model with HMT for long-context inputs. Since Stage 1 involves a shorter training sequence length and a simpler recurrent architecture than Stage 2, training with Stage 1 is faster per iteration (1.15 s/iteration) than Stage 2 (3.36 s/iteration). Within the same number of training steps, multi-stage training obtains better effectiveness and lower total training time than single-stage training. 

\section{HMT Memory Retrieval Behavior} \label{sec:mem_behavior}

One insight of using memory retrieval in HMT is handling frequent context switching to previously discussed topics or new topics. To evaluate this property, we employ PubMedQA and artifact the dataset with multiple contexts, mentioned in Section \ref{sec:res}. In this section, we will discuss other dataset manipulations on PG-19 to investigate the memory retrieval behavior of HMT further.

\begin{figure}[ht]
    \centering
    \begin{minipage}{0.45\textwidth}
        \includegraphics[width=0.9\textwidth]{ 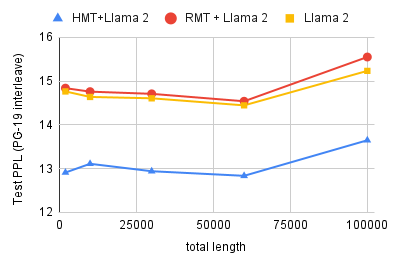}
        \caption{Effectiveness of HMT and RMT with Llama 2 7B evaluated over PG-19 with interleaving samples. HMT is 12.02\% better than RMT in terms of PPL for 2k to 100k-token samples.}
        \label{fig:interleave}
    \end{minipage} \hfill
    \begin{minipage}{0.45\textwidth}
        \includegraphics[width=0.9\textwidth]{ 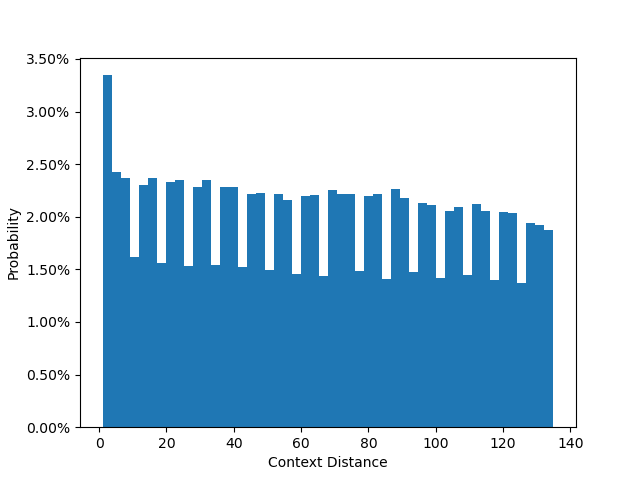}
        \caption{Histogram of context distance between the current segment and the segment corresponding to the memory embedding with the highest softmax score in the memory retrieval mechanism. The dataset evaluated is the PG-19 with interleaving samples.}
        \label{fig:interleave-context}
    \end{minipage}
\end{figure}

One way to manually introduce context switching is by interleaving the samples. For every 2 samples in the PG-19 dataset, we alternatively concatenate a segment of 256 tokens in each sample together to create a new sample. Therefore, a context switch will be invoked every 256 tokens. We fine-tuned and benchmarked HMT with Llama 2 7B over the artifact dataset. As a result, HMT can enhance the effectiveness of the baseline Llama 2 model, while RMT will worsen it, as shown in Figure \ref{fig:interleave}. We record the context distance of memory retrieval for 30k-token input, illustrated in Figure \ref{fig:interleave-context}, and notice a periodical recall distribution, indicating that HMT can capture the context-switching pattern. 

\begin{figure}[ht]
    \centering
    \includegraphics[width=0.5\columnwidth]{ 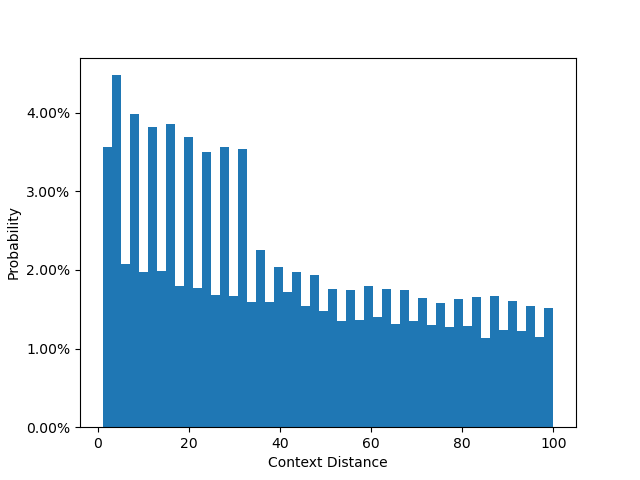}
    \caption{Histogram of context distance between the current segment and the segment corresponding to the memory embedding with the highest softmax score in the memory retrieval mechanism. The dataset evaluated is the dilated PG-19 dataset. Each sample is 25.6k tokens.}
    \label{fig:dilate}
\end{figure}

To demonstrate that HMT's behavior is aligned with the context-switching pattern, we further manipulate the PG-19 dataset by inserting 256 ``\$" tokens for every 256 tokens to dilate each sample. Intuitively, the segment containing ``\$" should be considered as irrelevant information and recalled infrequently. Figure \ref{fig:dilate} shows the memory retrieval pattern of HMT with Llama 2 7B over the dilated PG-19 dataset. We observe that HMT not only exhibits a periodical recall pattern but also successfully captures the position of irrelevant segments and avoids recalling them. 

\section{Relationships Between Effectiveness and Size of Sensory Memory}

During the experiment, we observed a general trend in the relationships between the effectiveness of HMT-augmented models and the size of sensory memory: the effectiveness will be first enhanced and then degraded as more and more embeddings are preserved for sensory memory. For instance, Table \ref{tab:sensory} illustrates the change of effectiveness of HMT + Llama 2 7B evaluated on Wikitext-103 with different sensory memory sizes. The PPL drops to the minimum when having 32 embeddings for the sensory memory.

\begin{table}[ht]
\caption{Effectiveness of HMT + Llama 2 7B evaluated on Wikitext-103 with 100k-token samples, with various sensory memory sizes. The segment size is 256 tokens. The effectiveness improves and then degrades with an increasing number of embeddings that are preserved for sensory memory.}
\label{tab:sensory}
\vskip 0.15in
\begin{center}
\begin{small}
\begin{sc}
\begin{tabular}{cc}
\toprule
\# of embeddings for sensory memory & Test PPL (Wikitext) \\
\midrule
8 & 4.54 \\
16 & 4.25 \\
32 & 4.19 \\
64 & 4.31 \\
128 & 4.57 \\
\bottomrule
\end{tabular}
\end{sc}
\end{small}
\end{center}
\vskip -0.1in
\end{table}

\section{Dataset Construction for PubMedQA} \label{sec:pubmedqa}

The original PubMedQA dataset does not have training, validation, and test dataset splits. In the experiments, we choose the \verb|pqa_artificial| subset and partition the training, validation, and test split, where the training split is the first 75\% of samples, the validation split is the next 15\% of samples, and the test split is the remaining 10\% samples. 

We artifact the long-context dataset from PubMedQA as the following: (1) select $M$ question-context-answer tuples from the dataset. Let this set of tuples be $\{(C_0, Q_0, A_0), (C_1, Q_1, A_1), \dots, (C_T, Q_T, A_T)\}$, where $C_n$ are contexts, $Q_n$ are questions, $A_n$ are answers for $0 \ge n \ge M$. Answers can be either long answers with detailed reasoning or short answers (either ``yes", ``no", or ``maybe"). (2) Concatenate all the contexts from each tuple to form a long context and append the questions and answers for each tuple. This will create $M$ sequences: ${C_0C_1\dots C_TQ_0A_0}$, ${C_0C_1\dots C_TQ_1A_1}, \dots$, ${C_0C_1\dots C_TQ_TA_T}$. By controlling the value of $M$, we can determine the fraction of useful information in the context for each question and better understand the filtering and selection ability of HMT and the baseline model.

\section{Gradient Stability in HMT and RMT} \label{sec:gradient}

\begin{figure}[ht]
    \centering
    \includegraphics[width=0.45\columnwidth]{ 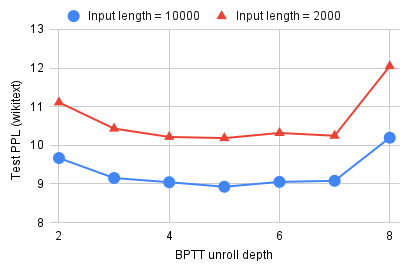}
    \caption{Effectiveness of training RMT with BPTT with different unroll depths for 2K tokens and 10K tokens input from the Wikitext-103 dataset. The backbone model is OPT 2.7B, with 256 tokens per segment during inference.}
    \label{fig:bptt}
\end{figure}

Both HMT and RMT are trained using backward propagation through time (BPTT) \cite{mozer2013focused}, a technique utilized to train the RNN model by unrolling recurrent forward passes of the model to optimize long-sequence learning. One issue with RMT training with BPTT is the gradient explosion and vanishing problem. With a higher BPTT unroll depth, the effectiveness of RMT will first increase and then decrease, with a slow reduction or even an increase in training loss. As seen in Figure \ref{fig:bptt}, we use the Wikitext-103 dataset with various BPTT unroll depths to access the effectiveness of RMT with the OPT 2.7B backbone model. For both 2k and 10k token inputs, we observe a rising PPL when unrolling more than 5 segments during training. 

\begin{table}[ht]
\caption{Relationship between the BPTT unroll depth and the test PPL of Wikitext-103 for OPT 2.7B with HMT. The experiment is evaluated on samples with 10k tokens. HMT preserved 32 tokens from the previous segment as the sensory memory and saved 300 memory embeddings for the memory retrieval. The segment size is 256 tokens.}
\label{tab:hmt_ppl}
\begin{center}
\begin{small}
\begin{sc}
\resizebox{0.5\columnwidth}{!}{
\begin{tabular}{cc}
\toprule
BPTT Unroll Depth & Test PPL (Wikitext) \\
\midrule
2 & 9.36 \\
5 & 9.15 \\
15 & 8.20 \\
\bottomrule
\end{tabular}
}
\end{sc}
\end{small}
\end{center}
\end{table}

Unlike RMT, HMT does not suffer from gradient vanishing or explosion as BPTT unroll depth increases due to the memory retrieval mechanism. Table \ref{tab:hmt_ppl} reveals that HMT can improve its effectiveness continuously as the BPTT unroll depth increases during training. Therefore, HMT will be more effective when the BPTT unroll depth increases. A detailed gradient stability analysis is presented in Appendix \ref{sec:gradient}. Furthermore, we applied several techniques to optimize the GPU memory consumption to increase the maximum trainable BPTT unroll depth compared with RMT, described in Appendix \ref{sec:dist}.


\begin{figure}[ht]
    \centering
    \includegraphics[width=0.8\columnwidth]{ 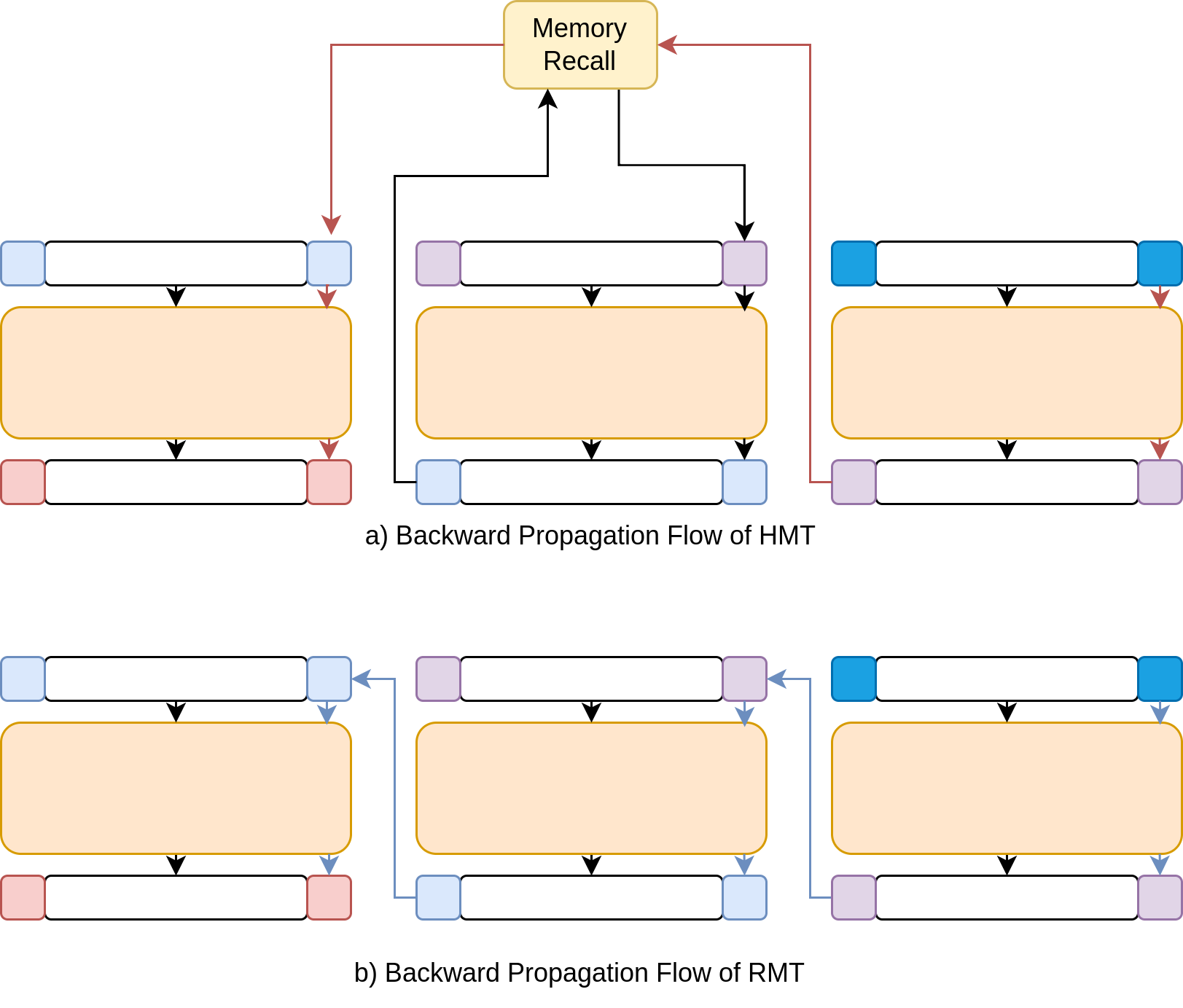}
    \caption{Backward propagation flows of HMT and RMT. The gradient of the first memorization prompt embedding (the red block on the right of the first segment) has multiple branches through the memory retrieval unit. Where the HMT gradient does not require propagation through each segment, the RMT gradient does.}
    \label{fig:gradient_hmt}
\end{figure}

In this section, we will formulate a mathematical description for the gradient stability of HMT when training with BPTT. BPTT with RMT behaves similarly to RNN which suffers from a vanishing or exploding gradient when there is a long chain of the gradient graph after unrolling \cite{pascanu2013difficulty}. Specifically, for a generic RNN model with the following form:
\begin{equation*}
    H_t = \sigma(\mathbf{A}H_{t-1} + \mathbf{B}x_t)
\end{equation*}
where $H_{t}$ is the hidden state at time $h$, $x_t$ is the input at time $t$, and $\mathbf{A}$ and $\mathbf{B}$ are parameters, the gradient will explode if $||A^T|| > 1$ and vice versa. A similar phenomenon occurs when training segment-level recurrent models such as RMT. Here we provide a scratch calculation on the gradient of loss with respect to the memory token at the starting time, which is one of the parameters in both RMT and HMT, after $t$ steps for RMT. Let $y'_{t+1} = H(x_t,m_t)$ be the logits and $m_{t+1} = F(x_t,m_t)$ be the generated memory embedding at time $t$, where $x_{t}$ is the input, $m_t$ is the memory token. The loss of inference is 
\begin{equation}
    L_{t+1} = \mathcal{L}(y'_{t+1}, y_{t+1})
\end{equation}
Therefore, the gradient can be calculated by the chain rule as
\begin{equation}
    \begin{split}
        \frac{\partial L_{t+1}}{\partial m_0} &= \frac{\partial L_{t+1}}{\partial y'_{t+1}} \times \frac{\partial y'_{t+1}}{\partial m_0} \\
        &= \frac{\partial L_{t+1}}{\partial y'_{t+1}} \times \frac{\partial H}{\partial m_t}(x_t) \times \frac{\partial m_t}{\partial m_0} \\
        &= \frac{\partial L_{t+1}}{\partial y'_{t+1}} \times \frac{\partial H}{\partial m_t}(x_t) \times \prod_{i=0}^{t-1} \frac{\partial F}{\partial m_i}(x_i)
    \end{split}
\end{equation}

Whether the gradient will explode or vanish depends on the input distribution and the function $F_t$. If $\forall x_t, \frac{\partial F}{\partial m_t}(x_t) > 0$, then the gradient will explode. Otherwise if $\forall x_t, \frac{\partial F}{\partial m_t}(x_t) < 0$, the gradient vanishes. Consequently, training RMT with a very high BPTT unroll depth can be inefficient. For HMT, with the assistance of the memory retrieval mechanism, the gradient is not prone to explosion or vanishing. Intuitively, the backward propagation of HMT for the memorization prompt embedding contains multiple short sub-branches to prevent gradient vanishing and the memory retrieval mechanism can modulate the propagation chain to avoid gradient explosion (Figure \ref{fig:gradient_hmt}). Let $G_t(z_t, m_{t}, m_{t-1}, \dots, m_{1}) = m'_t$ be the memory search function where $z_t$ is the representation extraction of segment at time $t$. Let $s$ be the summarization token for representation extraction. The gradient for HMT is
\begin{equation}
    \begin{split}
        \frac{\partial L_{t+1}}{\partial m_0} &= \frac{\partial L_{t+1}}{\partial y'_{t+1}} \times \frac{\partial y'_{t+1}}{\partial m_0} \\
        &=\frac{\partial L_{t+1}}{\partial y'_{t+1}} \times \frac{\partial H}{\partial m'_t}(x_t) \times \frac{\partial G_t}{\partial m_0} \\
        &= \frac{\partial L_{t+1}}{\partial y'_{t+1}} \times \frac{\partial H}{\partial m'_t}(x_t) \times (\sum_{k=1}^{t} \frac{\partial G_t}{\partial m_k}(z_k, m_{t}, \dots, m_{k-1}, m_{k+1}, \dots, m_1) \times \frac{\partial F}{\partial m_0}) \\
        &= \frac{\partial L_{t+1}}{\partial y'_{t+1}} \times \frac{\partial H}{\partial m'_t}(x_t) \times (\sum_{k=1}^{t} \frac{\partial G_t}{\partial m_k}(z_k, m_{t}, \dots, m_{k-1}, m_{k+1}, \dots, m_1) \times \frac{\partial F}{\partial m'_k} \times \frac{\partial G_{t-1}}{\partial m_0}) \\
        &= \dots \\
    \end{split}
\end{equation}
The root cause of the gradient explosion or vanishing comes from the long chain of gradient products in the formulation. For HMT, there are multiple short branches of the multiplication chain after expanding the expression. The longest chain over all components in the gradient is 
\begin{equation}
    \frac{\partial L_{t+1}}{\partial y'_{t+1}} \times \frac{\partial H}{\partial m'_t}(x_t) \times (\prod_{k=1}^{t-1} \frac{\partial F}{\partial m'_k} \times  \frac{\partial G_k}{\partial m_k}) \times \frac{\partial F}{\partial m_0}
\end{equation}

For gradient vanishing, since $\frac{\partial L_{t+1}}{\partial m_0}$ still has components with a shorter chain, the gradient will not disappear even when $||\frac{\partial F}{\partial m'_k} \times  \frac{\partial G_k}{\partial m_k}|| < 1$. For gradient explosion, empirically, $\frac{\partial G_k}{\partial m_{k}}$ are different for each $k$ by the property of cross attention and can modulate the term $\frac{\partial F}{\partial m_{k}}$ to distribute near 1. Thus, HMT is less prone to gradient explosion.

A similar proof can be deduced for the segment-level summarization token embedding of HMT for representation extraction.

\section{Distributed Training with Memory Consumption Optimization} \label{sec:dist}

Although \cite{bulatov2022recurrent} proves that unrolling more segments can improve the model effectiveness, they limit the number of segments unrolled to 4 with 2 NVIDIA A100 80GB GPUs since the maximum BPTT unroll depth is bounded by the GPU VRAM limit. There are three sources of VRAM consumption: model parameters, intermediate data (input segments, long-term memory, raw outputs of each segment, etc.), and optimization data (gradient and optimizer states). Although the computations of later segments do not require the intermediate data from the previous segment, the original BPTT will keep them on GPU by default. To reduce memory consumption, we customize the program to offload and load intermediate data for each input segment between the CPU and GPUs and distribute optimizer states and gradients throughout multiple GPUs running Zero Redundancy Optimizer (ZeRO) \cite{rajbhandari2020zero} Stage 2 in DeepSpeed \cite{rasley2020deepspeed}. These allow the model to unroll up to 15 segments with HMT. To train larger models, we employ LoRA \cite{hu2021lora} with rank 8. This allows us to fit 7B models to 4 MI210 GPUs.

\section{License and Links of Datasets and Models} \label{sec:license}

Datasets:

\begin{itemize}
    \item Wikitext (Texts from Wikipedia): CC BY-SA 4.0, \url{https://huggingface.co/datasets/Salesforce/wikitext}
    \item PG-19 (Books from Project Gutenberg): Apache License V2.0, \url{https://huggingface.co/datasets/emozilla/pg19}
    \item PubMedQA (Biomedical QA dataset): MIT License, \url{https://huggingface.co/datasets/qiaojin/PubMedQA}
    \item RedPajamaV2 (Pretraining data corpus): Apache License V2.0, \url{https://huggingface.co/datasets/togethercomputer/RedPajama-Data-V2}
    \item LongBench (Long context evaluation dataset): Licenses are mentioned in the original work \cite{bai2023longbench}, \url{https://huggingface.co/datasets/THUDM/LongBench}
\end{itemize}

\noindent Models:

\begin{itemize}
    \item OPT models: MIT License, \url{https://huggingface.co/facebook/opt-350m}
    \item Llama models: Llama 2 Custom License, \url{https://huggingface.co/meta-llama/Llama-2-7b-hf}
    \item OpenLlamaV2: Apache License V2.0, \url{https://huggingface.co/openlm-research/open_llama_3b_v2}
    \item RWKV: Apache License V2.0, \url{https://huggingface.co/RWKV/rwkv-4-3b-pile}
    \item Mamba: Apache License V2.0, \url{https://huggingface.co/state-spaces/mamba-370m-hf}
    \item Mistral: Apache License V2.0, \url{https://huggingface.co/mistralai/Mistral-7B-v0.3}
    \item Yi: Yi-License, \url{https://huggingface.co/01-ai/Yi-6B-200K}
    \item Qwen: Tongyi Qianwen License, \url{https://huggingface.co/Qwen/Qwen1.5-0.5B}, \url{https://huggingface.co/Qwen/Qwen2.5-14B}
    \item SmolLM: Apache License V2.0, \url{https://huggingface.co/HuggingFaceTB/SmolLM-135M}
\end{itemize}

\noindent All datasets and models are publicly available and free to use.

\end{document}